\documentclass{article}

\usepackage{arxiv}

\usepackage[utf8]{inputenc} % allow utf-8 input
\usepackage[T1]{fontenc}    % use 8-bit T1 fonts
\usepackage{hyperref}       % hyperlinks
\usepackage{url}            % simple URL typesetting
\usepackage{booktabs}       % professional-quality tables
\usepackage{amsfonts}       % blackboard math symbols
\usepackage{amsmath}
\usepackage{nicefrac}       % compact symbols for 1/2, etc.
\usepackage{microtype}      % microtypography
\usepackage{lipsum}		% Can be removed after putting your text content
\usepackage{graphicx}
\usepackage{natbib}
\usepackage{doi}
\usepackage{authblk}   % Required for multiple affiliations
\usepackage[nameinlink,capitalise]{cleveref}
\usepackage{makecell}   % newline in one cell
\usepackage{subcaption}
\usepackage{soul,xcolor}

\newcommand{\shorttablewidth}{0.3\linewidth}

\title{Neonatal Hypoxic-ischaemic Encephalopathy Classification from the EEG and HRV Signals Using a Conformer based Masked Autoencoder
	\footnotesize \textsuperscript{}
	\thanks{The work was supported by Taighde Éireann – Research Ireland (19/FFP/6782). This study was also supported by a Strategic Translational Award and an Innovator Award from the Wellcome Trust (098983 \& 209325).}
}

\date{} 					% Or removing it

\author[1,2]{Shuwen Yu\thanks{Corresponding author: shuwenyu@umail.ucc.ie}\hspace{0.4em}}
\author[1,2]{Shuwen Yu\hspace{0.4em}}
\author[1,2]{William P Marnane}
\author[2,3]{Geraldine B. Boylan}
\author[1,2]{Gordon Lightbody}

\affil[1]{Department of Electrical \& Electronic Engineering, School of Engineering and Architecture\\ University College Cork, Cork, Ireland}
\affil[2]{INFANT Research Centre, University College Cork, Cork, Ireland}
\affil[3]{Pediatrics and Child Health, University College Cork, Cork, Ireland}
\renewcommand{\headeright}{}
\renewcommand{\shorttitle}{MAEConformer: Neonatal HIE Classification from EEG and HRV Signals}

\renewcommand{\th}{\textsuperscript{th}} % ex: I won 4\th place

\hypersetup{
pdftitle={Neonatal Hypoxic-ischaemic Encephalopathy Classification from the EEG and HRV Signals Using a Conformer based Masked Autoencoder},
pdfsubject={cs.LG, cs.AI, eess.SP},
pdfauthor={Shuwen Yu, William P.~Marnane, Geraldine B.~Boylan, Gordon Lightbody},
pdfkeywords={Masked Autoencoders, Conformer, Electroencephalography, Heart Rate Variability},
}

\begin{document}
\maketitle

\begin{abstract}
In this paper, we propose the MAEConformer, a novel self-supervised learning framework that combines the Conformer architecture with the Masked Autoencoder (MAE) paradigm for large-scale representation learning from unlabelled electroencephalography (EEG) and heart rate variability (HRV) signals. By integrating convolutional operations with Transformer-based self-attention, MAEConformer effectively captures both local temporal patterns and long-range contextual dependencies in physiological time series. To enhance reconstruction fidelity and representation quality, a multi-resolution short-time Fourier transform (MR-STFT) loss is incorporated alongside the reconstruction objective, enabling the model to jointly learn temporal and spectral characteristics across multiple scales. Modality-specific EEG and HRV MAEConformer models were pretrained on 6,030\,h and 4,868\,h of unlabelled recordings, respectively, and subsequently transferred to expert-annotated downstream tasks. Experimental results demonstrate that the learned representations provide strong transferability and data efficiency. In EEG-based hypoxic–ischemic encephalopathy (HIE) severity classification, the pretrained MAE-EEG model achieved test AUCs of 97.19\% and 96.56\% for binary and four-class classification tasks, respectively, outperforming a range of state-of-the-art supervised and self-supervised baselines. On the HRV-based HIE severity classification task, MAE-HRV achieved a test AUC of 82.42\%, surpassing both self-supervised Transformer-based and supervised convolutional baselines. These findings demonstrate the effectiveness of MAEConformer for learning robust and transferable representations across multiple physiological modalities. The code is publicly available at \url{https://github.com/syu-kylin/MAEConformer}.
\end{abstract}

% keywords can be removed

\section{Introduction}
Neonatal hypoxic-ischemic encephalopathy (HIE) is a brain injury caused by oxygen deprivation and reduced cerebral blood flow around the time of birth \cite{allenHypoxicIschemicEncephalopathy2011,hankinsDefiningPathogenesisPathophysiology2003}. Despite advances in neonatal care, HIE remains a major cause of mortality and long-term neurological disability. Therapeutic hypothermia (TH) is currently the standard treatment for infants with moderate-to-severe HIE; however, its effectiveness depends on initiation within the first six hours after birth \cite{kosteczkoTherapeuticHypothermiaForm2025}. Consequently, accurate and timely assessment of HIE severity is critical for clinical decision-making and outcome prediction.

Electroencephalography (EEG) is widely regarded as the gold standard for monitoring neonatal brain injury in Neonatal Intensive Care Units (NICUs). However, continuous EEG monitoring requires specialized equipment and expert interpretation, resources that are not always available on a 24/7 basis \cite{chockOptimalNeuromonitoringTechniques2023}. Heart rate variability (HRV), while less sensitive than EEG, has emerged as a promising non-invasive biomarker of autonomic and neurological function and is routinely available in most NICUs \cite{bersaniHeartRateVariability2021}. Together, EEG and HRV provide complementary information that may facilitate earlier, more objective, and continuous assessment of neonatal encephalopathy severity.

The increasing availability of physiological recordings has driven the development of automated machine learning systems for neonatal monitoring. Deep learning approaches have demonstrated superior performance across a wide range of biomedical applications, yet their success typically relies on large quantities of expertly annotated data \cite{del2023applications,wu2024neuro}. In neonatal neurocritical care, obtaining such annotations is costly, time-consuming, and inherently difficult to scale, creating a major barrier to the broader adoption of deep learning methods \cite{rani2024self}.

Self-supervised learning (SSL) addresses this challenge by learning informative representations directly from unlabelled data. Among SSL approaches, Masked Autoencoders (MAEs) have emerged as an effective framework for representation learning, where a model learns to reconstruct masked portions of the input from the remaining visible observations \cite{krishnan2022self}. After pretraining, the learned encoder can be transferred to downstream tasks using only a limited amount of labelled data. This paradigm has demonstrated strong performance in computer vision and is increasingly being adopted for physiological signal analysis\cite{zhouEnhancingRepresentationLearning2024, zhouMaskedTransformerElectrocardiogram2024}. 

Despite their success, existing MAE frameworks present two limitations when applied to physiological time series. First, most approaches rely on pure Transformer backbones, which lack the strong local inductive biases required to effectively model short-term physiological dynamics. Although convolutional modules can complement self-attention by capturing local patterns, their integration into MAE frameworks is non-trivial because standard convolutions operate on dense inputs and may introduce masking information leakage \cite{wooConvNeXtV2Codesigning2023,gaoConvMAEMaskedConvolution2022,zhang2023hivit}. Second, reconstruction objectives are typically dominated by time-domain losses such as mean squared error (MSE), which may not adequately preserve important spectral characteristics of physiological signals \cite{enganFHRFormerSelfsupervisedTransformer2025, zhang2023self}.

To address these limitations, we propose the MAEConformer, a self-supervised learning framework that integrates the Conformer architecture with the Masked Autoencoder paradigm for large-scale pretraining on unlabelled EEG and HRV recordings. By combining convolutional operations with Transformer-based self-attention, MAEConformer captures both local temporal structures and long-range contextual dependencies while avoiding masking information leakage through the use of visible-token encoding. In addition, a multi-resolution short-time Fourier transform (MR-STFT) loss is incorporated to complement the reconstruction objective and encourage the learning of richer spectral representations across multiple temporal scales.

The main contribution of this work are summarized as follows: 1). A unified self-supervised framework, MAEConformer, was proposed, which combined the strengths of Conformer architectures and Masked Autoencoders for physiological time-series representation learning. 2). A frequency-aware reconstruction objective based on MR-STFT loss was introduced to improve spectral representation learning during pretraining. 3). the modality-specific MAEConformer models were pretrained on 6,030\,h of EEG recordings and 4,868\,h of HRV recordings, respectively, and evaluate their transferability on downstream HIE severity classification tasks. 4). Extensive experiments demonstrate that MAEConformer learns robust and transferable representations across both EEG and HRV modalities. The pretrained MAE-EEG model achieves state-of-the-art performance on binary and multi-class EEG-based HIE severity classification, while MAE-HRV consistently outperforms existing self-supervised and supervised HRV baselines.

The remainder of this paper is organized as follows: Section~2 reviews related work, Section~3 presents the proposed methodology, Section~4 reports the experimental results and ablation studies, and Section~5 discusses the findings and concludes the paper.

\section{Related Work}
\subsection{Masked Autoencoders for Physiological Signals}
The Masked Autoencoder (MAE) framework has recently been extended from computer vision to physiological signal analysis. Several studies have demonstrated the effectiveness of self-supervised masked reconstruction for learning transferable biosignal representations. For example, \cite{zhouMaskedTransformerElectrocardiogram2024} employed a Transformer-based MAE for ECG representation learning, while \cite{wangUnsupervisedPreTrainingUsing2023} adopted a ConvNeXt V2 backbone for self-supervised ECG pretraining. In the EEG domain, \cite{zhouEnhancingRepresentationLearning2024} proposed a multichannel MAE that simultaneously masks temporal segments and EEG channels, using a cosine-similarity objective to guide reconstruction. Unlike BENDR \cite{kostasBENDRUsingTransformers2021}, which performs masked prediction in the latent feature space, MAEEG \cite{chienMAEEGMaskedAutoencoder2022} reconstructs the original EEG signal through additional linear and convolutional decoding layers and optimizes a reconstruction objective directly in the signal domain.

Among existing approaches, MAE-EEG-Transformer \cite{caiMAEEEGTransformerTransformerbasedApproach2024} and FHRFormer \cite{enganFHRFormerSelfsupervisedTransformer2025} are most closely related to the present work. MAE-EEG-Transformer employs a standard Transformer encoder-decoder architecture and reconstructs masked EEG segments using an MSE loss, whereas FHRFormer combines time-domain reconstruction with a frequency-domain focal frequency loss for fetal heart rate reconstruction. These studies demonstrate the potential of MAE-based pretraining for physiological signals, but primarily rely on Transformer backbones and relatively short input sequences.

\subsection{Convolution and Frequency-Aware Learning in MAEs}
Although Transformer-based MAEs have shown strong representation learning capability, several studies have explored integrating convolutional operations to enhance local feature modelling. However, incorporating convolution into MAE frameworks is non-trivial because standard convolutions operate on dense inputs and may introduce masking information leakage. To address this challenge, ConvNeXt V2 \cite{wooConvNeXtV2Codesigning2023} and ConvMAE \cite{gaoConvMAEMaskedConvolution2022} employ specialized sparse or masked convolutional operations, while HiViT \cite{zhang2023hivit} adopts a hierarchical architecture to maintain efficient masked modelling. Despite these advances, the effective integration of convolutional inductive biases and masked reconstruction remains an open problem for physiological time-series representation learning.

Another line of research seeks to improve reconstruction quality through frequency-domain supervision. Motivated by the frequency principle \cite{xuTrainingBehaviorDeep2019,xuFrequencyPrincipleFourier2020}, several studies have incorporated frequency-aware objectives alongside conventional reconstruction losses. Examples include focal frequency loss \cite{jiangFocalFrequencyLoss2021}, frequency-domain Transformer architectures \cite{piaoFredformerFrequencyDebiased2024}, and frequency-aware masked autoencoders \cite{liuFrequencyAwareMaskedAutoencoders2024}. While these approaches improve spectral reconstruction, most rely on Fourier-domain objectives that provide limited temporal localization. For highly non-stationary physiological signals such as EEG and HRV, preserving both temporal and spectral information is critical.

In contrast to previous studies, the proposed MAEConformer combines a Conformer backbone with the MAE framework to jointly model local and long-range temporal dependencies while avoiding masking information leakage. Furthermore, a multi-resolution short-time Fourier transform (MR-STFT) loss is employed to capture time-varying spectral characteristics across multiple temporal scales, making the framework particularly suitable for long-duration physiological recordings.

\section{Methods}
\subsection{Datasets}
This study is a secondary data analysis from two multicentre cohort studies where data were collected from newborns recruited from January 2011 to February 2017 across eight European tertiary neonatal intensive care units from four European countries (Ireland, the Netherlands, Sweden and the UK). The original studies aim to assess the feasibility and effectiveness of cEEG monitoring on seizure and accuracy of Algorithm of Neonatal Seizure Recognition (ANSeR). Infants born at or after 36+0 weeks of gestation who required EEG monitoring for clinical reasons were included. Ethical approval was granted for both studies by national and local Ethics Committees specific to each participating centre. The primary findings have been published, and detailed study protocols are available on \href{https://clinicaltrials.gov/}{ClinicalTrials.gov} (Identifiers: NCT02160171 \cite{rennieCharacterisationNeonatalSeizures2019}, NCT02431780 \cite{pavelMachinelearningAlgorithmNeonatal2020}).

Among 472 enrolled neonates, 284 were diagnosed with hypoxic-ischemic encephalopathy (HIE) and had at least 6\,h of EEG recordings. After excluding infants with combined diagnoses, recordings commencing later than 48\,h after birth, and a separate hold-out cohort reserved for future validation, 181 neonates remained for this study, comprising 91 recordings from ANSeR1 and 90 recordings from ANSeR2. Continuous EEG was acquired using either the Nihon Kohden Neurofax EEG-1200 or the NicoletOne ICU Monitor/Xltek systems at sampling frequencies of 250 or 256\,Hz. ECG channels were available for 58 infants in ANSeR1 and 75 infants in ANSeR2.

For each infant, up to five one-hour epochs (approximately 6, 12, 24, 36 and 48\,h after birth) with minimal artefacts were selected and graded by expert neurophysiologists according to the scheme of \cite{murrayEarlyEEGFindings2009}. This produced 338 and 315 strongly labelled EEG epochs from ANSeR1 and ANSeR2, respectively. HRV signals were extracted from the corresponding ECG recordings using an enhanced Pan-Tompkins algorithm \cite{yu2026hrvconformer}, yielding 215 and 257 labelled HRV epochs after quality control. In addition, the remaining unannotated recordings from ANSeR2 were used for self-supervised pretraining, producing approximately 6,030 one-hour EEG epochs and 4,868 one-hour HRV epochs.

The unlabelled dataset was extracted from all the remaining recordings for each baby. The earliest recordings began approximately within the first hour after birth, while the latest extended up to 228 hours postnatally. As the ANSeR1 dataset will be used for testing, only the ANSeR2 data was involved for training. As a result, the ANSeR2 dataset produced approximately 6,030 and 4,868 one-hour epochs of unannotated EEG and HRV recordings, which are used for pretraining. 

The expert annotated each one-hour epoch of EEG grade into five classes (from 0 to 4): \textit{normal}, \textit{mild abnormality}, \textit{moderate abnormality}, \textit{major abnormality} and \textit{inactive}. Treatments are needed if diagnosed as moderate to severe (grade 2 to 4) --- usually with \textit{therapeutic hypothermia} (TH). For binary classification, we therefore categorized the normal and mild abnormality as class 0 (TH not required), while the more severe grades (moderate, severe and inactive) are then grouped as class 1 (TH required). Alternatively, for a more elaborate and fine-grained classification, the normal and mild abnormality are combined as class 0, which produces 4 classes to keep consistent with prior studies \cite{raurale2021grading, o2023development}.

\subsection{Pre-processing}
For the EEG signal, the default reference montage was first converted to bipolar montage using the available 8 channels: F4-C4, C4-O2/C4-P4, F3-C3, C3-O1/C3-P3, T4-C4, C4-Cz, Cz-C3, and C3-T3. Then, these signals were  band-pass filtered between 0.5 and 13 Hz using a finite impulse response (FIR) filter to retain the most relevant frequencies. The 0.5\,Hz cut-off frequency aims to attenuate the very low frequency artefacts, such as DC drift. As these signals were recorded in different centres and machines even with different protocols, the sampling frequency changes either with 256\,Hz or 250\,Hz. All of them were uniformly downsampled to 16\,Hz using FFT-based resampling with anti-aliasing for further processing. 

Each one-hour epoch was split into 5-minute windows with 50\% overlap to feed into the model and capture some relevant features. For the strong label data, each 5-minute window shares the same label as its one-hour epoch. As HIE is a global injury, channels are treated independently and concatenated to formulate one-dimensional input, which not only increases the number of training samples, but also allows training a channel-variant EEG classifier. The resultant model can therefore make decisions even with a single channel, preparing for the adaptation to different downstream tasks. As the signal mean is already zero (the DC drift was removed by the band-pass filter), a scaling factor of $1 \times 10^4$ simply scales the signal to the range of [-1, 1] approximately, thereby improving numerical stability and optimization efficiency during neural network training.

For the HRV signal, RR intervals were extracted from the ECG recordings using the enhanced Pan-Tompkins algorithm and processed with the same procedures described in \cite{yu2026hrvconformer}. Specifically, each one-hour epoch was split into separate long segments if an irregular RR interval (e.g. very long intervals) were detected. For instance, if there is a RR interval over a certain value (for example 4\,s), this one-hour epoch would be then split into two segments effectively removing this irregular RR interval. As the RR intervals are extracted from the ECG signal, they are irregularly sampled. For the convenience of further processing, uniform resampling is necessary to produce the NN intervals. The resampling includes a two-step interpolation --- the signal was first linearly interpolated with a native ECG sampling frequency (256\,Hz) to obtain a dense representation, which improve the robustness of the subsequent cubic spline interpolation in absence of irregular RR intervals or missing beats \cite{cliffordQuantifyingErrorsSpectral2005}. This was followed by a cubic spline interpolation with a sampling frequency of\,4 Hz \cite{vestOpenSourceBenchmarked2018}, as the most relevant frequencies in HRV are below 2\,Hz \cite{gouldingHeartRateVariability2015}.

Each one-hour epoch (may include multiple long segments) was then split into 5-minute windows with 80\% overlap to produce more samples. They inherit the same label as their one-hour epoch (for the strong label group). Another noise removal step was subsequently applied to exclude any 5-minute window NN interval whose standard deviation exceeded a predefined threshold (e.g., 0.12\,s, empirically selected to balance signal quality and data retention), as such windows were considered excessively noisy and unsuitable for further analysis. Any one-hour epoch with less than 10 valid windows was excluded. Before being fed into the model, all 5-minute samples were normalized with min-max normalization to keep them in the range of [0, 1]. The minimum and maximum value was selected as the 5\th{} and 95\th{} percentiles of the dataset to avoid the effect of outliers.

\subsection{MAEConformer}
\begin{figure}[htbp]
	\centering
	\includegraphics[width=0.9\textwidth]{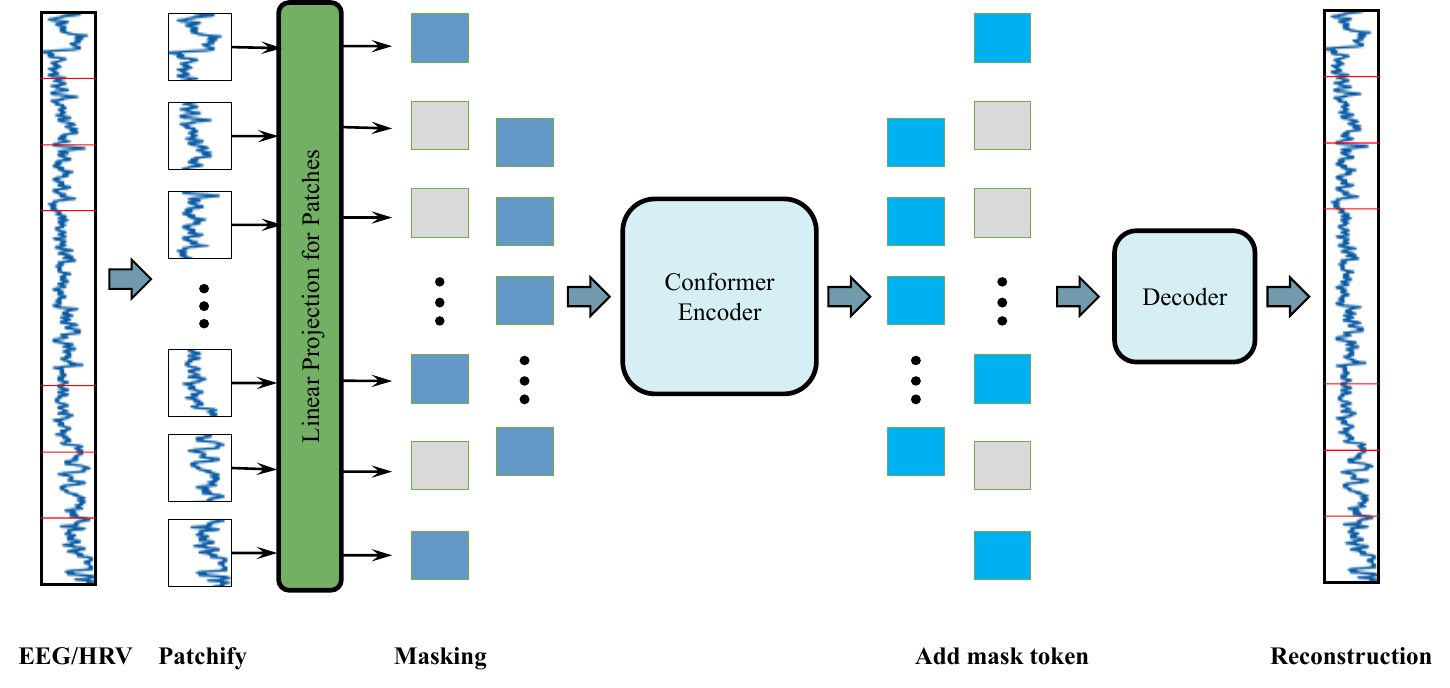}
	\caption[The model architecture of MAEConformer.]{The model architecture of MAEConformer. The input is a 5-minute HRV or EEG (single channel) segment. After being patchified, they are projected into a sequence of patch embeddings. A random masking removes some of the patches, and only the rest of the visible patches are sent into the Conformer encoder to learn some latent representation. The output of the encoder is added with a series of learnable masked tokens as placeholders for these missing patches. After restoring the sequence, they are fed into the decoder to reconstruct the missing patches.}
	\label{fig:maeconformer}
\end{figure}

The MAEConformer adopted a masked autoencoder architecture that reconstructs the original signal given the visible portion. The encoder maps the observed signal into a latent representation, and the decoder reconstructs the raw input from the latent representation together with mask tokens. As the decoder is only used in the pretrain stage, it allows an asymmetric design for the autoencoder. Like most of the MAE frameworks, a lightweight decoder is employed to enable highly efficient computation. The overview of the MAEConformer architecture is shown in \cref{fig:maeconformer}.

%%%%%%%%%%%%%%%%%%%% fig:MAEConformer %%%%%%%%%%%%%%%%%%%%%%%%

The input $x \in \mathbb{R}^{N}$ is a 5-minute 1-D HRV or EEG signal with $N$ samples. Like the ViT, the input sequence was divided into non-overlapped patches with patch size of $p_s$, producing a sequence of patches $x^p \in \mathbb{R}^{L \times p_s}$ ($L=N/p_s$). These patches were linearly projected into a latent space to learn richer representations with an embedding dimension of $d_{\mathrm{model} 
}$. Before being processed by the encoder, a random masking is implemented, which randomly samples a subset of patch embeddings without replacement and masks (i.e., discards) them. Only the remaining patches are sent to the encoder. Different masking strategies have been explored, such as block masking \cite{gaoConvMAEMaskedConvolution2022}, uniform masking \cite{liUniformMaskingEnabling2022}, and semantic-based masking \cite{liSemMAESemanticGuidedMasking2022, mohammadifoumaniEEG2RepEnhancingSelfsupervised2024, kakogeorgiouWhatHideYour2022}. Despite this, the random masking strategy is widely used because of its simplicity and effectiveness. It has been shown that random masking works well for audio \cite{yadavMaskedAutoencodersMultiWindow2024}, EEG and ECG signals \cite{zhouEnhancingRepresentationLearning2024, zhouMaskedTransformerElectrocardiogram2024}.

Unlike most MAE frameworks that employ a Vision Transformer (ViT) backbone, the proposed MAEConformer adopts the Conformer architecture \cite{gulatiConformerConvolutionaugmentedTransformer2020}, which combines Transformer-based self-attention for modelling long-range dependencies with convolutional modules for local feature refinement. This hybrid architecture has previously demonstrated strong performance on physiological time-series analysis, including HRV classification \cite{yu2026hrvconformer}. More importantly, it integrates naturally with the MAE framework. Rather than introducing specialized sparse or masked convolutions to preserve masked token locations, as in ConvNeXt V2 and ConvMAE, MAEConformer simply discards masked patches and processes only the visible tokens within the encoder. Furthermore, because the Conformer maintains a constant feature dimension throughout the encoder, it avoids the need for complex masking and reconstruction strategies, such as latent-space masking and feature upsampling. Consequently, the proposed design preserves the simplicity and computational efficiency of the original MAE framework while avoiding masking information leakage.

Rather than adding a fixed position embedding before masking, the relative position embedding was used in the attention layers of the Conformer block. Compared with the fixed position embedding, the relative position embedding can better generalize different sequence lengths and patch sizes. Unlike absolute position embedding which introduces non-uniform jumps in the encoder (after masked) leading to a soft form of masking leakage, the relative position embedding only operates on the compressed sequence without the notion of mask indices, avoiding the leakage of mask pattern or span length. Despite a distortion of temporal distance, the depthwise convolution operates on a compressed latent space, where neighbouring tokens may not be corresponding to adjacent patches in the original signal. That is analogous to the attention operation on non-continuous tokens, but constrained within a local region. In addition, the encoder is optimized to learn a robust representation from the partial observations rather than exacting timing representations, while any neighbourhood distortion present in the pretraining disappears in the downstream fine-tune when no masking is applied. 

It has been shown that simple MAE pretraining can suffer from feature collapse -- that is the learned embeddings are located in a low-dimensional subspace leading to redundant representations and poor performance \cite{wooConvNeXtV2Codesigning2023, zhangHowMaskMatters2022}. Solutions have been explored to improve the feature diversity from the activation and loss function perspectives. ConvNeXt V2 \cite{wooConvNeXtV2Codesigning2023} proposed a \textit{Global Response Normalization (GRN)} to suppress overly dominant channels and enhance weaker ones to encourage the feature competition resulting in diversity representations. 

Given an input feature of $X \in \mathbb{R}^{L \times d_{\mathrm{model}}}$, GRN is formulated by \cref{eqn:grn}, where $\mathcal{G}(x)$ is the aggregated features with $L_2$ norm across patches and $\mu$ is the mean of $\mathcal{G}(X)$. With scaling of features by the global response to the mean of all channels, this normalization introduce mutual inhibition that prevents energy concentration and encourages boarder channel participation in training.

\begin{equation}
	\label{eqn:grn}
	\begin{split}
		\mathcal{G}(X) &= \left \{ {\lVert X_{:,1}\rVert}_2, \dots, {\lVert X_{:,d_{\mathrm{model}}} \rVert}_2 \right \} \in \mathbb{R}^{d_{\mathrm{model}}} \\
		Y &= \gamma \odot (X \odot \frac{\mathcal{G}(X)}{\mu}) + \beta + X
	\end{split}
\end{equation}
with $\gamma, \beta \in \mathbb{R}^{d_{\mathrm{model}}}$ broadcast across the sequence dimension. Unlike batch normalization which requires batch-level statistics, GRN is more lightweight sample-level normalization. Importantly, it avoids the distribution shift across datasets between training and testing. Consequently, the batch normalization in the Convolution module was replaced with GRN to improve feature diversity and representation robustness. 

The input of the decoder is a full set of tokens, comprising the visible latent representations from the encoder and a set of mask tokens. The mask tokens are represented by a shared learnable vector, which serve as placeholders to tell the decoder that these patches need to be predicted. The decoder only relies on the mask token's position information and interaction between visible patches to recover the missing part. After being concatenated and reordered properly, this full set is processed by a stack of Conformer blocks. Similarly, relative position embedding is added into each token in the attention module. A linear head is attached to the Conformer blocks to reconstruct the masked signal. As the decoder is only used in the pretraining stage, it allows an independent design for the encoder. Like most of the MAE works, a lightweight narrower and shallower decoder is employed, enabling an efficient large-scale pretraining \cite{wangMaskedImageModeling2023, xieSimMIMSimpleFramework2022, feichtenhoferMaskedAutoencodersSpatiotemporal2022}.

\subsection{Loss Functions}
The MAE reconstruction target is the input normalized EEG or HRV signal. The decoder output is a sequence of vectors from the final linear head. Each of them represents the recovered single patch samples. After being reshaped, the prediction has the same shape as the input. The \textit{Mean Square Error (MSE)} function is employed as the main reconstruction loss as shown in \cref{eqn:mse_loss}. The MAE loss is only calculated on the masked patches, by marking the masked samples ($m_i$) as 1 and visible ones as 0 (assume $M$ samples are masked in the input). In addition to the MSE loss, another \textit{Uniformity loss} and \textit{Multi-resolution Short-Time Fourier Transform (MR-STFT)} loss are used as auxiliary loss. 

\begin{equation}
	\label{eqn:mse_loss}
	\mathcal{L}_{MSE} = \frac{1}{M} \sum_{i=1}^{N} (y_i - \hat{y}_i)^2 \cdot m_i
\end{equation}

\subsubsection{Uniformity loss}
The \textit{uniformity loss} was proposed in \cite{zhangHowMaskMatters2022} to improve the pretrained MAE feature diversity. It argues that MAE can drift the representation to a subspace where variance concentrates on a few directions, i.e., feature collapse. The uniformity loss works as a regularizer to explicitly penalize high similarity between features of unrelated samples, and encourage embeddings to be uniformly spread. 

Formally, it is defined in \cref{eqn:uniformity_loss}, where $x_u$ is an unmasked view of the input sample, while $x^-_u$ is another random draw of an unmasked sample. $g$ denotes the transformation function of the encoder. Most commonly they are normalized by the $L_2$ norm to encourage features to be spread on the unit hypersphere -- \textit{cosine similarity}. By leveraging the average square similarity between representation of a sample and another independent sample, it encourages the encoder to learn diversity embeddings. 

\begin{equation}
	\label{eqn:uniformity_loss}
	\mathcal{L}_{unif} = \mathbb{E}_{x_u} \mathbb{E}_{x^-_u}(g(x_u)^Tg(x^-_u))^2
\end{equation}

\subsubsection{Mask-aware Multi-resolution Short-Time Fourier Transform loss}
As MAE reconstructs the signal in the time domain, the MSE loss only compares the sample-by-sample amplitudes. Consequently, it ignores how the errors distribute in the time-frequency space. For example, two signals with similar spectra but shifting in time may cause low perceptual difference but high MSE loss, while signals with similar time trend but different frequency balance produce low MSE but large frequency error. Therefore, the \textit{Mask-aware Multi-resolution Short-Time Fourier Transform (MR-STFT)} loss is proposed to complement the MSE loss and improve the frequency representation.

The \textit{Short-Time Fourier Transform (STFT)} employs a moving window to describe how a signal frequency evolves over time. Given a specific resolution with window length of $L_i$ (FFT size of $N_i=L_i$), hop size of $H$, the discrete STFT at time frame $\ell$ using window function of $w[n]$ is defined as: 

\begin{equation}
	\label{eqn:stft}
	X^{(i)}[\ell,k]
	=
	\sum_{n=0}^{L_i-1}
	x[n+\ell H]\; w_i[n]\; e^{-j 2\pi k n / N_i},
	\qquad k=0,1,\dots,N_i-1.
\end{equation}

The commonly used \textit{Hann} window function is employed. The window size is an important factor to trade off the temporal and frequency resolution. A longer window produces a finer frequency resolution while reduces the temporal resolution. By combining multiple window lengths, it helps the model learn multi-resolution time-frequency representations.

As the STFT frames are generated using overlapping windows, many frames contain both masked and unmasked samples. Naively discarding all partially masked frames from the spectral loss would result in the loss of valuable boundary information near the masked regions. To address this issue, a mask-aware soft weight factor $w_{ma}$ was introduced. This weight quantifies the proportion of masked samples contributing to each frame, thereby enabling partial supervision of frames that overlap masked regions. In this manner, the reconstruction objective emphasizes masked regions while avoiding excessive penalization of unmasked signal components. Importantly, the same Hann window function used in the STFT computation is also applied when calculating the mask-aware weights. This ensures that the weighting mechanism precisely reflects the actual sample contributions to each spectral frame, maintaining consistency between the spectral transformation and the loss.

For a single frame within the resolution window $i$, the STFT loss $\mathcal{L}^{(i)}_{\mathrm{STFT}}(\ell)$ on each time frame $\ell$ is defined in \cref{eqn:lstft_frame}, where both the amplitude and complex differences are computed, and $\beta_c$ controls the contribution of complex-domain difference. The $\lVert \cdot \rVert_1$ denotes element-wise complex $L1$ norm over all frequency bins $k$, and $\left| \cdot \right|$ denotes the modulus of complex-valued STFT coefficients. Instead of explicitly constraining the phase discrepancy, the complex difference implicitly penalizes the phase error by measuring the Euclidean distance in the complex domain, resulting in a smoother and more stable optimization process. This formulation avoids several practical issues associated with direct phase constraints. For example, phase $\phi$ become ill-defined when the amplitude approaches zero; in contrast, the complex difference naturally down-weights phase contributions at low amplitudes. Moreover, as phase is discrete and periodic, explicit phase losses can introduce gradient discontinuities. In addition, even a small temporal shift $\epsilon$ can introduce a large phase error, given by $\Delta \phi = 2\pi f \epsilon$, particularly for the high-frequency components. The STFT loss for a single  resolution is a weighted average of all frames, expressed in \cref{eqn:lstft_resolution} where $\varepsilon$ is a small constant for numerical stability.

\begin{equation}
	\label{eqn:lstft_frame}
	\mathcal{L}^{(i)}_{\mathrm{STFT}}(\ell)
	=
	\left\lVert \,
	\left|X^{(i)}_{\ell}(\hat{x})\right|
	-
	\left|X^{(i)}_{\ell}(x)\right|
	\right\rVert_{1}
	\;+\;
	\beta_c
	\left\lVert \,
	X^{(i)}_{\ell}(\hat{x})
	-
	X^{(i)}_{\ell}(x)
	\right\rVert_{1}
\end{equation}

\begin{equation}
	\label{eqn:lstft_resolution}
	\mathcal{L}^{(i)}_{\mathrm{STFT}}
	=
	\frac{\sum_{\ell=0}^{T_i-1} w_{\mathrm{ma}}^{(i)}[\ell]\; \mathcal{L}^{(i)}_{\mathrm{STFT}}(\ell)}
	{\sum_{\ell=0}^{T_i-1} w_{\mathrm{ma}}^{(i)}[\ell] + \varepsilon}
\end{equation}

The MR-STFT loss is averaged over all frames and frequency bins across all resolution windows $n_i$, as shown in \cref{eqn:mrstft_loss}. A hyperparameter -- \textit{minimum masked weight} $\delta$ is added to discard any frames whose $w_{ma} \le \delta$ to avoid the influence of too many unmasked patches.

\begin{equation}
	\label{eqn:mrstft_loss}
	\mathcal{L}_{\mathrm{MRSTFT}}
	= \frac{1}{n_i}\sum_{i=1}^{n_i} \mathcal{L}^{(i)}_{\mathrm{STFT}}
\end{equation}

The final loss is the weighted sum of the MSE loss, uniformity loss and the MR-STFT loss, as shown in \cref{eqn:overall_loss}, where $\alpha_{\mathrm{unif}}$ and $\alpha_{\mathrm{MRSTFT}}$ are hyperparameters to weight the uniformity and MR-STFT loss, respectively.
\begin{equation}
	\label{eqn:overall_loss}
	\mathcal{L}
	=
	\mathcal{L}_{\mathrm{MSE}}
	+
	\alpha_{\mathrm{unif}}\,\mathcal{L}_{\mathrm{unif}}
	+
	\alpha_{\mathrm{MRSTFT}}\,\mathcal{L}_{\mathrm{MRSTFT}} .
\end{equation}

\subsection{Post-processing}
The model processes on the 5-minute sample-level, but the final prediction is reported on the one-hour epoch level. A post-processing stage was implemented to obtain performance metrics on the epoch-level. The epoch-level accuracy was aggregated by a majority vote method, which takes the most frequent predictions on the samples as the final one-hour epoch prediction. This is consistent with the expert annotation process, where HIE severity may change over time and the final decision is based on an overall assessment of the one-hour epoch.

In order to calculate the AUC over the one-hour epoch, a score needs to be aggregated from all the 5-minute window predictions from that epoch. The \textit{margin mean} or \textit{log-odds} aggregation method was implemented. The model's output $o_c$, before softmax, i.e., the logits, are used to calculate the margin. For a binary classification, the logit margin $m_w$ for each sample $w$ prediction can be defined as:

\begin{equation}
	m_w = o_1 - o_0
\end{equation}
\noindent where $o_1$ and $o_0$ are the predicted logits of the positive and negative class respectively. This is equivalent to the logit-odd of class 1 versus class 0 probability: 

\begin{equation}
	sample\_margin = log \frac{p(y=1 | x)}{p(y=0 | x)}
\end{equation}

A larger margin indicates a more confident prediction, which is consistent with the rank property of ROC-AUC. Assuming that each one-hour epoch has $n$ samples, and they were predicted independently, the one-hour epoch prediction probability ($log$) can be represented as:

\begin{equation}
	log \frac{P(y=1 | epoch)}{P(y=0 | epoch)} \propto \sum_{i=0}^{n-1}{log \frac{p(y=1|x_i)}{p(y=0 | x_i)}}
\end{equation} 

In this way, the sum of the samples margin in one epoch can represent the probability of that epoch prediction. To reduce noise, the sample margins are averaged as the prediction score for that epoch. However, this method is only suitable for the AUC calculation of binary classification. For the multi-class AUC calculation, the \textit{one-vs-rest} (ovr) strategy is adapted, which fits the classifiers from one class against all the other classes. For instance, in the 4-class model, each sample has 4 logits, producing multiple margins. It is not meaningful to pick any of them and therefore losing structure for class-wise AUC calculation. Consequently, multi-class AUC score epoch aggregation in this work employed the average of per-class probabilities.

\section{Results}

\subsection{Experimental setup}
The pretrain was performed on large unlabelled EEG (6030\,h) and HRV (4868\,h) datasets respectively collected from ANSeR2. Around 5\% of each dataset was separated out as a validation set. They were both pretrained with the same MAEConformer architecture, but slightly different configurations. The model was trained using the AdamW optimizer ($\beta_1=0.95, \beta_2=0.99$). The cosine warmup learning rate scheduler was adopted with 50 epochs of warmup period and a learning rate of $5\times 10^{-5}$. The training was implemented with Pytorch Distributed Data Parallel (DDP) run on a SLURM system with a request of 2 NVIDIA L40S GPUs (46 GB). Under the default configuration, pretraining the MAE-HRV model for 1500 epochs required approximately 3.86 hours, while pretraining the MAE-EEG model for 500 epochs required approximately 8.32 hours due to the larger number of channels and higher sampling frequency of the EEG signals.

The downstream linear probe and fine-tune were trained on the ANSeR2 strong label group (315\,h EEG and 257\,h HRV respectively). 20\% of this data (63\,h EEG and 51\,h HRV) was used as a validation set for model selection. The best model is selected with the criterion of the highest moving average validation AUC. Model performance was tested on independent unseen EEG (338\,h) and HRV (215\,h) strong label groups respectively provided from ANSeR1. The linear probe used the pretrained encoder attached with a linear classifier head (a batch norm layer and a linear layer). Only the linear head was optimized, while the pretrained encoder was frozen. No regularization (dropout and weight decay) was used during linear probing. The AdamW optimizer was used with a learning rate of $5\times 10^{-5}$. The fine-tune model was initialized with pretrained weights and an additional fully convolutional network (FCN) classifier head. For fine-tuning, both the pretrained encoder and classifier head were optimized. The AdamW optimizer parameters were set as $\beta_1=0.9, \beta_2=0.999$. Layer wise learning rate decay (decay rate 0.85) was used with base learning rate of $8\times 10^{-5}$. The code is available at \url{https://github.com/syu-kylin/MAEConformer}.

Model hyperparameters were optimized with Optuna \cite{optuna_2019} based on downstream validation set using a KNN classifier, followed by a finer optimization with linear probe. The final configurations with key hyperparameters of both models are shown in \cref{tab:mae_model_configs}.

% %%%%%%%%%%%%%%% table of model configurations %%%%%%%%%%%%%%%%%%%%%%%%%
\begin{table}[htb]
	\caption{MAEConformer model configurations.}
	\centering
	\begin{tabular}{ll}
		\toprule
		models & configuration \\
		\midrule
		% MAE-EEG
		MAEConformer (EEG) & 
		\makecell[l]{
			patch size: 10\,s \\
			embedding dimension: 256  \\
			number of layers: 6 \\
			number of attention heads: 16 \\
			depthwise conv kernel size: 7 \\
			decoder embed dimension: 128 \\
			decoder depth: 3 \\ 
			decoder attention heads: 8 \\
			mask ratio: 0.4 \\
			position embedding: relative \\
			classifier head: FCN \\
			fcn head conv kernel size: 7 \\
			uniformity loss weight: 0.2 \\
			mrstft loss weight: 0.6 \\
			stft window length: (4, 8, 16, 32)s \\
			min mask weight: 0.4 \\
		} \\
		\midrule
		
		% MAE-HRV
		MAEConformer (HRV) &
		\makecell[l]{
			patch size: 10\,s \\
			embedding dimension: 256  \\
			number of layers: 6 \\
			number of attention heads: 16 \\
			depthwise conv kernel size: 7 \\
			decoder embed dimension: 128 \\
			decoder depth: 3 \\ 
			decoder attention heads: 8 \\
			mask ratio: 0.3 \\
			position embedding: relative \\
			classifier head: FCN \\
			fcn head conv kernel size: 7 \\
			uniformity loss weight: 0.3 \\
			mrstft loss weight: 0.3 \\
			stft window length: (4, 8, 16, 32, 64)s \\
			min mask weight: 0.4 \\
		} \\
		\bottomrule
	\end{tabular}
	\label{tab:mae_model_configs}
\end{table}

\subsection{Evaluation}
\begin{figure}[htbp]
	\centering
	\includegraphics[width=0.45\textwidth]{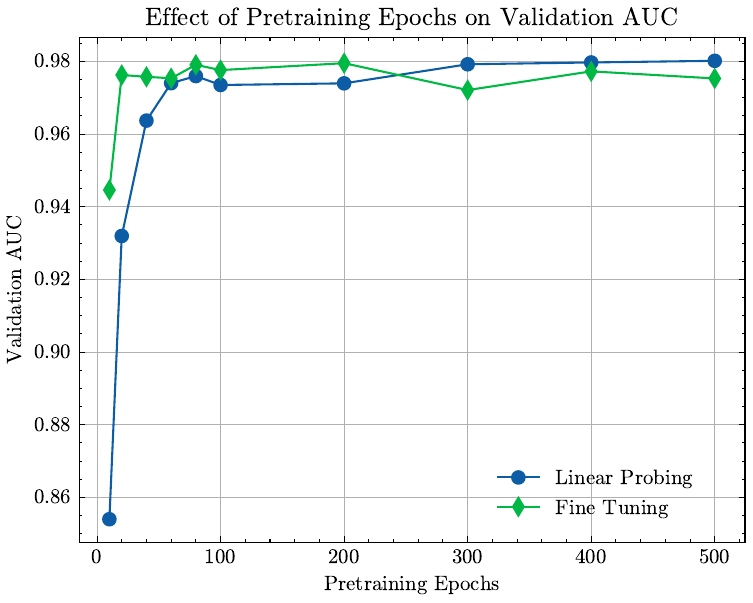}
	\caption{The effect of the number of pretrain epochs on the downstream classification performance from the 4-class (average AUC on the epoch-level) EEG validation set.}
	\label{fig:eeg_pretrain_epoch_val_auc}
\end{figure}

The MAEConformer was pretrained on unlabelled EEG and HRV datasets for 500 and 1500 epochs respectively, until no obvious performance improvement observed over the validation set. Due to its much lower number of available channels, the MAE-HRV requires a longer epoch pretrain. The effect of the number of pretrained epochs on the downstream classification performance over the EEG dataset is demonstrated in \cref{fig:eeg_pretrain_epoch_val_auc}. With 100 pretrain epochs, the classification performance on the linear probe and fine-tune achieved a significant improvement. After 300 epochs pretrain, the validation AUC on the linear probe almost plateaued, while the fine-tune performance peaked by 200 epochs. 

%%%%%%%%%%%%%%%% Figure: EEG pretrained epoch vs performance %%%%%%%%%%%%%%%%%%%%

\begin{figure}[htb]
	\centering
	\begin{subfigure}[b]{0.8\textwidth}
		\centering
		\includegraphics[width=\textwidth]{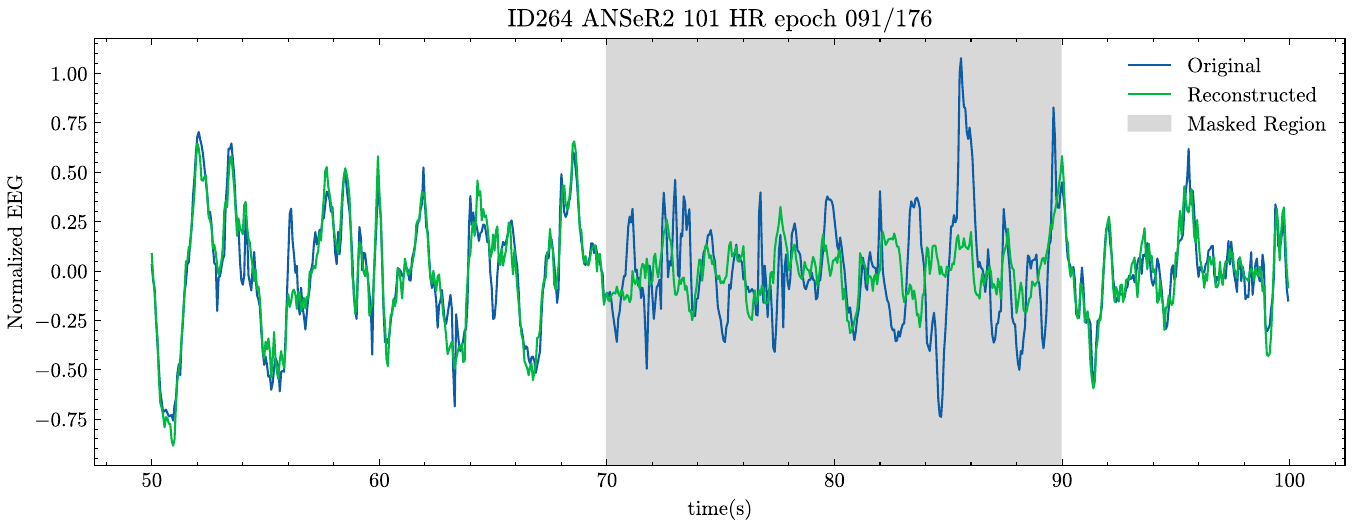}
		\caption{}
		\label{subfig:eeg_reconstruct_example_1}
	\end{subfigure}
	\hfill
	\begin{subfigure}[b]{0.8\textwidth}
		\centering
		\includegraphics[width=\textwidth]{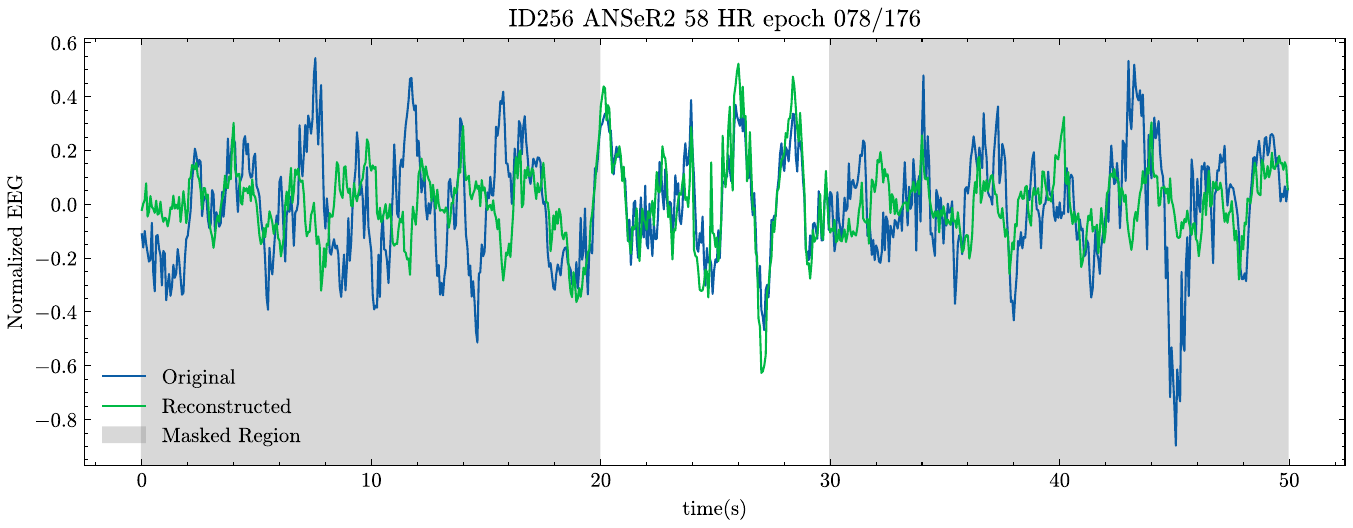}
		\caption{}
		\label{subfig:eeg_reconstruct_example_2}
	\end{subfigure}
	\caption{Examples of MAE-EEG reconstructions from the validation set. 50 seconds of EEG segments were randomly selected from 5-min window (to allow for more detailed visualization). Both the mask ratio and STFT window min mask weight were set as 0.4 and patch size of 10\,s. Although the reconstructed EEG were not perfectly overlapped with the raw signal, the general trend has been captured in the mask region.}
	\label{fig:eeg_reconstruct_examples}
\end{figure}

\begin{figure}[htb]
	\centering
	\begin{subfigure}[b]{0.8\textwidth}
		\centering
		\includegraphics[width=\textwidth]{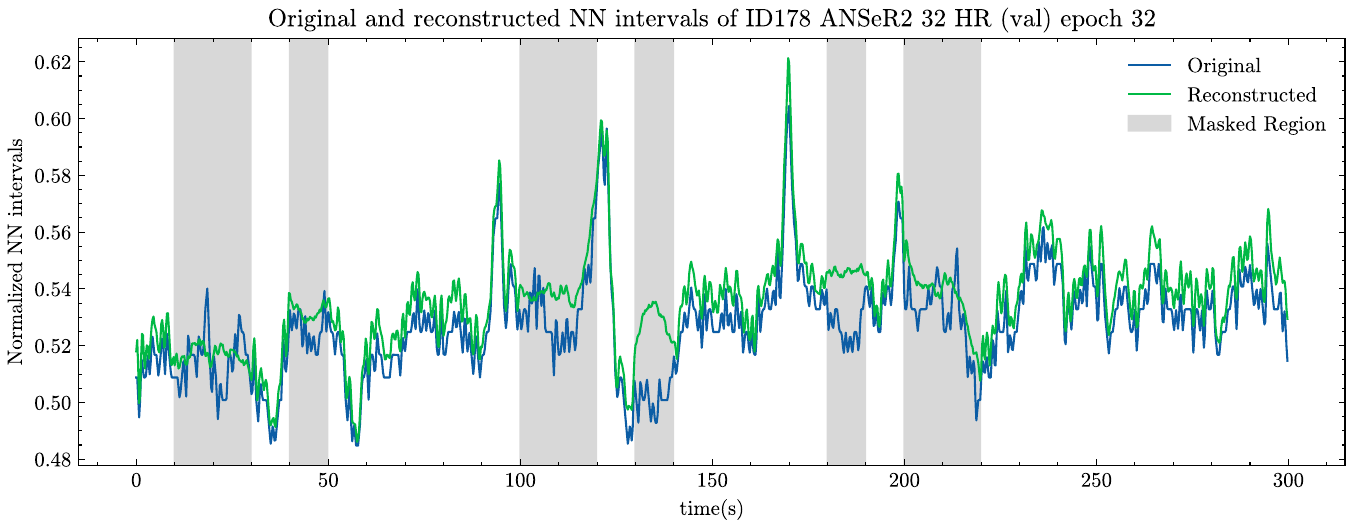}
		\caption{}
		\label{subfig:hrv_reconstruct_example_1}
	\end{subfigure}
	\hfill
	\begin{subfigure}[b]{0.8\textwidth}
		\centering
		\includegraphics[width=\textwidth]{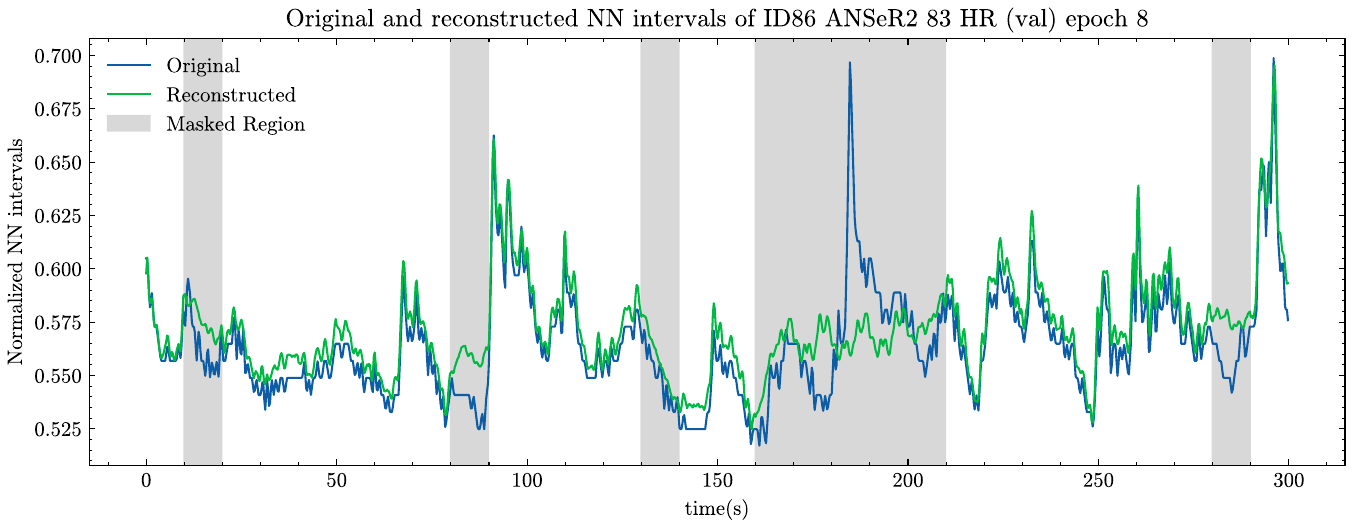}
		\caption{}
		\label{subfig:hrv_reconstruct_example_2}
	\end{subfigure}
	\caption{Examples of reconstructed HRV signal from the validation set. The MAE-HRV (mask ratio of 0.3, patch size of 10s) accurately reconstructs the NN intervals in the unmasked patches, while in the masked area it recovers the overall temporal trends without generating artificial spike.}
	\label{fig:hrv_reconstruct_examples}
\end{figure}

Some examples of the reconstructed EEG and HRV signal are exhibited in \cref{fig:eeg_reconstruct_examples} and \cref{fig:hrv_reconstruct_examples}. Signals are visualized using pretrained validation sets to evaluate the models' generalization capability. The MAEConformer used the configurations listed in \cref{tab:mae_model_configs}. To clearly show the EEG details, only part of segments (50\,s) from 5-minute windows are shown, while the whole 5-minute window for HRV signal is shown due to its lower frequency content. It can be seen that both MAE models can recover the raw signal nearly perfectly in the unmasked region, indicating effective identification preservation. However, in the masked patches the reconstructed signals are not completely overlapped with the original signal, but the overall temporal trend has been captured without inducing spurious artefacts, which suggests the reconstruction emphasizes physiologically plausible dynamics rather than overfitting to noise. 

%%%%%%%%%%%%%%%% Figure: EEG reconstruction example %%%%%%%%%%%%%%%%%%%%

%%%%%%%%%%%%%%%% Figure: hrv reconstruction example %%%%%%%%%%%%%%%%%%%%

\subsubsection{Partial fine-tune}
Compared with the EEG dataset, the strongly labelled HRV dataset contains substantially fewer recordings, making full-network fine-tuning more susceptible to overfitting. In preliminary experiments, full fine-tuning consistently underperformed linear probing. Therefore, a partial fine-tuning strategy was investigated, whereby only the top $n$ encoder layers were unfrozen while the remaining layers were kept fixed.

\begin{figure}[htb]
	\centering
	\includegraphics[width=0.45\textwidth]{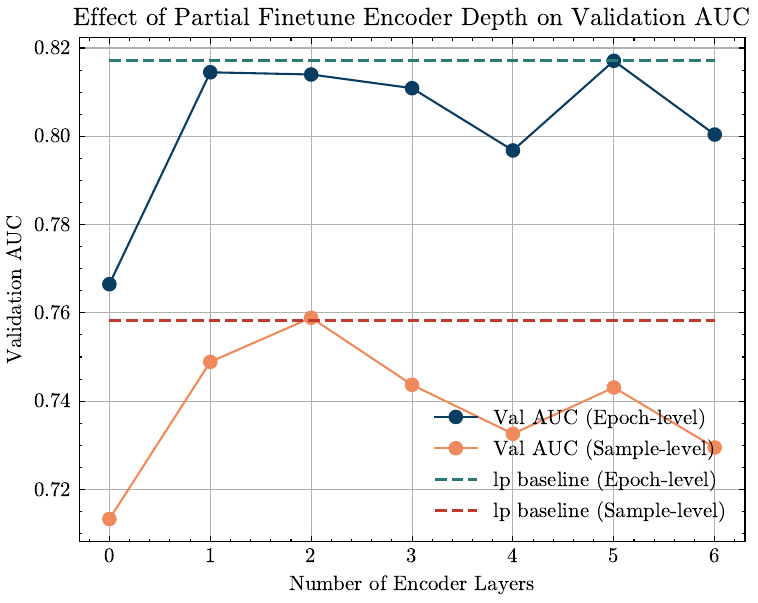}
	\caption{Effect of partial fine-tune with the top $n$ layers of MAE-HRV encoder on validation performance. Fine-tune encoder layer 0 indicates only the classifier head is optimized (default with FCN head). Line probe (lp) baselines on the sample-level and epoch-level are marked with dash lines.}
	\label{fig:hrv_partical_ft_depth}
\end{figure}

The results are shown in \cref{fig:hrv_partical_ft_depth}. Three independently pretrained models were fine-tuned and evaluated using the same training and validation sets. The linear probing performance remained constant averaged across all experiments and is included as a reference. Fine-tuning only the default FCN classifier head (top 0 layers) resulted in substantially lower performance than linear probing, primarily due to the reduced capacity of the FCN head compared with the linear probe classifier. Unfreezing the top two encoder layers significantly improved performance, achieving validation AUCs comparable to linear probing at both the sample and epoch levels. Further unfreezing of lower encoder layers generally led to a decline in validation performance, indicating that updating a larger number of parameters was not beneficial under the limited-label setting. Based on these results, fine-tuning the top two encoder layers was adopted as the default strategy for MAE-HRV.
%%%%%%%%%%%%%%%% Figure: hrv partical ft encoder depth %%%%%%%%%%%%%%%%%%%%

\subsubsection{Pretrain effect}
\begin{figure}[htbp]
	\centering
	\includegraphics[width=0.45\textwidth]{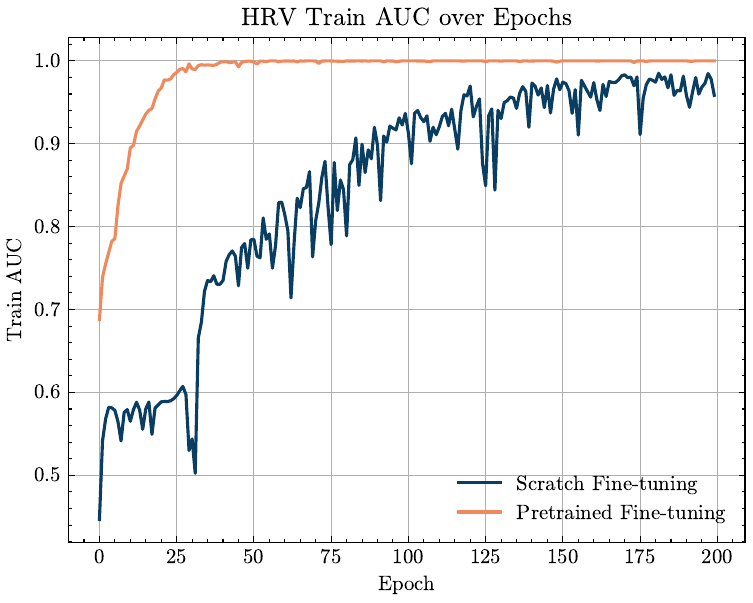}
	\caption{The downstream task training efficiency comparison between models whose encoder were initialized randomly (scratch fine-tune) and with pretrained weights (pretrained fine-tune). Both models were optimized across the whole network (including the encoder and FCN classifier head) on downstream HRV strongly labelled dataset.}
	\label{fig:hrv_train_efficiency}
\end{figure}

\cref{fig:hrv_train_efficiency} demonstrates the pretrain effect from the perspective of downstream task training efficiency. A randomly initialized Conformer encoder with a fully convolutional network (FCN) classifier head was jointly optimized to conduct an HRV classification task. This network starting from a random guess (around 0.5 of AUC) to nearly perfect classification (0.98 of AUC over training set) takes around 200 epochs. In contrast, model loading with pretrained weights has a higher starting point -- 0.7 of AUC, which suggests that the pretrained model has learned meaningful representations. Moreover, this pretrained network with only 25 epochs optimization can achieve close to 100\% of training AUC, which provides approximate 8-times acceleration. In addition, compared with the randomly initialized network, the training curve using the pretrained weights is smoother, which indicates a more stable optimization process.

%%%%%%%%%%%%%%%% Figure: hrv training efficiency %%%%%%%%%%%%%%%%%%%%

\begin{figure}[htbp]
	\centering
	\includegraphics[width=0.45\textwidth]{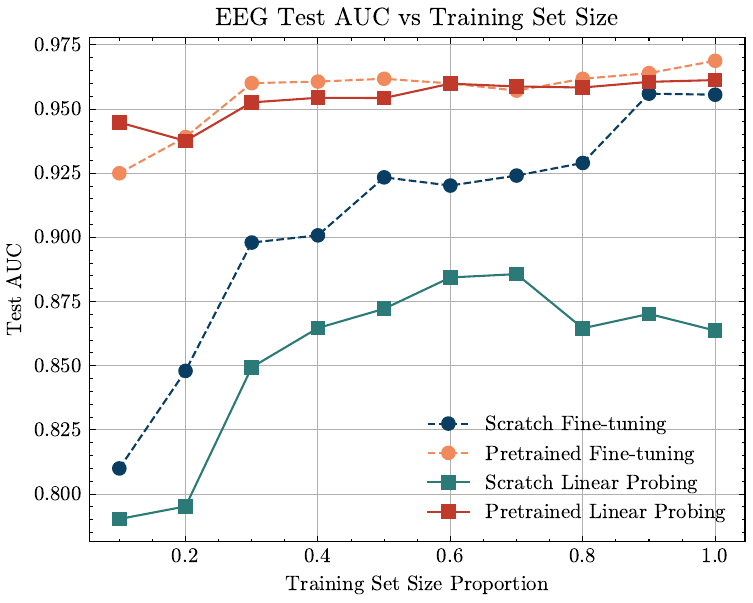}
	\caption{The effect of different size of strongly labelled EEG training set on the test AUC (4-class on the epoch-level). Model training from scratch can achieve similar performance as the model loaded from pretrain only with full size of training data (252 one-hour epochs).}
	\label{fig:eeg_test_auc_vs_train_size}
\end{figure}

To better demonstrate the advantage of the pretrained network, the scaling behaviour of MAEConformer was investigated. The pretrained MAE-EEG and MAE-HRV models were fine-tuned using different fractions of the training set from the EEG and HRV datasets respectively. As a baseline, models with identical structure but randomly initialized weights (train from scratch) were evaluated under the same settings. \cref{fig:eeg_test_auc_vs_train_size} presents the epoch-level test AUC (average over 4 classes) for models trained on varying proportions of the 4-class EEG datasets. Overall, models initialized with pretrained weights exhibit strong robustness to training set size. Only modest improvement is observed, when the proportion of training set increase from 10\% to 30\% (linear probe: 92.5\% to 96\%; fine-tune: 94.5\% to 95.5\%). Beyond this point, the linear probe and fine-tune performance stabilize at approximately 96\%-97\%. Furthermore, the minimal gap between linear probe and fine-tune suggests that the pretrained encoder has learned a highly stable and linearly separable representation. 

%%%%%%%%%%%%%%%% Figure: eeg train data scaling %%%%%%%%%%%%%%%%%%%%

In contrast, model training from scratch demonstrates a steep performance increasing as the training set grows, indicating meaningful feature representations are gradually learned with increasing data availability. When training with the full dataset, the scratch fine-tune model achieved comparable performance to the pretrained model. However, the performance of scratch linear probe remains substantially constrained as expected given randomly initialized encoder. Interestingly, the linear probe shows moderate improvement as the training proportion increasing from 10\% to 70\% (test AUC rising from 79\% to 88\%), suggesting some degree of linear separability may emerge from random Conformer features under sufficient data. Nevertheless, the large gap between training from scratch and pretrained fine-tune under limited data (81\% vs. 92.5\% with only 10\% of training data) clearly highlights the advantage of pretraining, particularly in low-resource scenarios.

A similar scaling analysis was conducted for MAE-HRV model. Due to comparatively smaller dataset size, performance variability across validation splits was more pronounced. Therefore, the test performance was averaged over five random train/validation splits. In each split, the validation set is fixed (20\% strongly labelled ANSeR2 data), while the remaining data were proportionally allocated for training. During fine-tune from the pretrained model, only the top 2 encoder layers and the classifier head were optimized, with earlier layers frozen.

%%%%%%%%%%%%%%%% Figure: hrv train data scaling %%%%%%%%%%%%%%%%%%%%
\begin{figure}[htbp]
	\centering
	\includegraphics[width=0.45\textwidth]{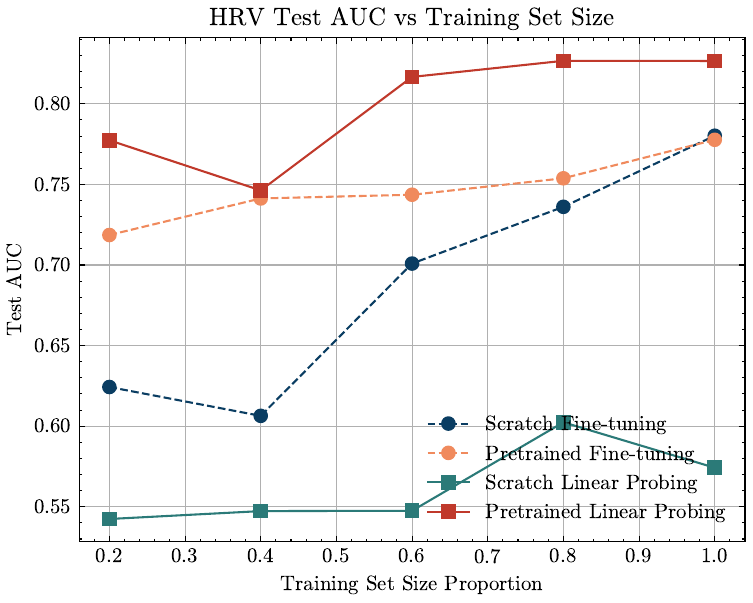}
	\caption[The effect of different size of HRV training set on the model performance]{The effect of different size of the strongly labelled HRV training set on the test AUC (2-class on the epoch-level). The full size of training data includes 206 one-hour epochs while the validation set keep constant as 51 epochs. Each point is averaged over five random train/val splits evaluation.}
	\label{fig:hrv_test_auc_vs_train_size}
\end{figure}

The results are shown in \cref{fig:hrv_test_auc_vs_train_size}. Again, pretrained models consistently outperform their scratch counterparts across different percentages of training set sizes. The pretrained fine-tune model shows a gradual improvement from 72\% to 78\% test AUC, as the training proportion increases from 20\% to 100\%. In contrast, the scratch fine-tune model starts at a lower baseline (63\%) but rise sharply to 78\% with more data, underscoring the critical role of training data when no prior representation is available. A frozen random encoder with learnable linear head performs near chance level regardless of the training size, serving as a reference to quantify the contribution of the pretrained encoder. In contrast, linear probe with pretrained encoder demonstrates strong data efficiency. Even with only 20\% of training data, the model achieves 78\% test AUC. Although a slight dip is observed at 40\% of data, it recovers and reaches 81\% with 60\% of the training data, after which, it stabilizes. These findings strongly confirm the effectiveness of the pretrain process and its ability to produce robust and data-efficient representations.

\subsection{Comparison with baselines}
\paragraph{\underline{Baseline architectures:}}
To demonstrate the capability of the proposed method, the downstream classification performance is compared against several baselines. Here, two self-supervised architectures and several fully supervised learning methods are selected and evaluated on the same test set. The FHRFormer, originally proposed in \cite{enganFHRFormerSelfsupervisedTransformer2025}, employs a MAE architecture to reconstruct the fetal heart rate signal. This model employs a standard Transformer encoder, while the decoder introduces an extra cross attention sublayer in the Transformer block. In addition to the mean squared error (MSE) loss, it also includes an amplitude-based focal frequency loss as auxiliary loss. This method is adapted as a benchmark, termed MAEFHRFormer, and employs the same architectural configurations as MAEConformer. 

Another baseline is adapted from the MAE-EEG-Transformer \cite{caiMAEEEGTransformerTransformerbasedApproach2024}, which was originally pretrained on BCI EEG datasets for motor imagery classification. The original model employed a Transformer encoder and a lightweight MLP decoder, as it operates on comparatively short EEG segments. Reconstruction is optimized solely using the MSE loss. However, the present task requires the model to process substantially longer 5-minute EEG windows, which significantly increase temporal complexity compared to the original setting. Under this configuration, the original MAE-EEG-Transformer is unable to effectively reconstruct the raw signal. To address this limitation, the MLP decoder was replaced with a standard Transformer decoder to enhance temporal modelling capacity. In addition, the same focal frequency loss was incorporated alongside the MSE loss to better preserve spectral characteristics and stabilize long-sequence reconstruction. The resulting adapted architecture is referred to as MAETransformer. For fair comparison, it adopts the same architectural configurations (e.g., embedding dimension, depth and patch size) as MAEConformer.

\paragraph{\underline{EEG classification task:}}
For EEG classification, three convolutional neural networks were selected as benchmarks as they have reported the performance on the same test set. The first one employed the fully convolutional network (FCN) directly processing on the raw EEG signal \cite{yuNeonatalHypoxicIschemicEncephalopathy2023}. The other two, using a time-frequency distribution feature map as input, employed a CNN architecture with  either a fully connected layer (TFD-CNN) classifier head \cite{raurale2021grading} or linear regression  (TFD-CNN-Reg) \cite{o2023development} head.

The test performance of different models evaluated on the ANSeR1 4-class EEG dataset are presented in \cref{tab:maeeg_baseline_4_class}. These three MAE models' performance are averaged over six independent experiments with different random initializations. From the results, the MAEConformer consistently outperforms the other five baselines across all metrics, except for the fine-tune accuracy which was slightly inferior to the FCN model. However, the MCC of MAEConformer is superior to the FCN model. In terms of self-supervised learning, the MAEConformer achieved a substantial margin over the other two MAE baselines, particularly in the linear probe performance, highlighting the advantage of the proposed method. 

Interestingly, the lightweight FCN model achieved a similar test AUC as MAEFHRFormer and even higher MCC and accuracy than the both MAE baselines, underscoring the effectiveness of the FCN architecture for this particular EEG classification task. The MAEFHRFormer has a narrow lead across all metrics than the MAETransformer from both linear probe and fine-tune performance. This difference arise from optimization-induced effects rather than the encoder architectural changes, highlighting the role of decoder design in shaping encoder representations during the mask autoencoder pretraining.

%%%%%%%%%%%% table of EEG (4-class) baseline comparison %%%%%%%%%%%%%%%%%%%%%%%%%
\begin{table}[htbp]
	\caption{Comparison of different models' performance over the EEG test set (one-hour epoch-level) with 4-class prediction.}
	\centering
	\begin{tabular}{l|ccc|cc}
		\toprule
		& \multicolumn{3}{c|}{fine-tune} & \multicolumn{2}{c}{linear probe} \\
		& MCC      &    ACC   &   AUC   &    ACC          &     AUC       \\ 
		\midrule
		MAEConformer   & $\mathbf{0.7781_{\pm0.027}}$ & $0.8639_{\pm0.016}$ & $\mathbf{0.9656_{\pm0.003}}$ & $\mathbf{0.8481_{\pm0.007}}$ & $\mathbf{0.9630_{\pm0.001}}$    \\
		MAEFHRFormer   & $0.6686_{\pm0.013}$ & $0.8013_{\pm0.008}$ & $0.9382_{\pm0.006}$ & $0.7071_{\pm0.003}$ & $0.8790_{\pm0.009}$ \\ 
		MAETransformer & $0.6189_{\pm0.006}$ & $0.7722_{\pm0.034}$ & $0.9274_{\pm0.016}$ & $0.6509_{\pm0.047}$ & $0.8504_{\pm0.021}$ \\  \hline
		FCN \cite{yuNeonatalHypoxicIschemicEncephalopathy2023}     & 0.7691   & $\mathbf{0.869}$   & 0.9328 &     -           &       -       \\
		TFD-CNN \cite{raurale2021grading}       & 0.722    & 0.828   &   -     &     -           &       -       \\
		TFD-CNN-Reg \cite{o2023development}   &    -     & 0.695   &   -     &     -           &       -       \\ 
		\bottomrule
	\end{tabular}
	\label{tab:maeeg_baseline_4_class}
\end{table}

These three pretrained MAE models were subsequently fine-tuned on the same EEG dataset but using binary (2-class) labels. As expected, performance metrics improved to a varying extent due to reduced classification complexity. Importantly, this improvement indicates that the effectiveness of the pretrained representations is not dependent on the specific label granularity of the downstream task. Across all evaluation metrics, MAEConformer consistently maintains the leading position, further demonstrating the robustness and generalizability of its learned representations.

%%%%%%%%%%%%%%% table of EEG (2-class) baseline comparison %%%%%%%%%%%%%%%%%%%%%%%%%
\begin{table}[htbp]
	\caption{Comparison of different MAE models performance on the EEG test set (epoch-level) with 2-class prediction.}
	\centering
	\begin{tabular}{l|ccc|cc}
		\toprule
		& \multicolumn{3}{c|}{fine-tune} & \multicolumn{2}{c}{linear probe} \\
		& MCC      &    ACC   &   AUC   &    ACC          &     AUC       \\ 
		\midrule
		MAEConformer   & $\mathbf{0.8181_{\pm0.029}}$ & $\mathbf{0.9068_{\pm0.018}}$ & $\mathbf{0.9719_{\pm0.001}}$ & $\mathbf{0.9053_{\pm0.005}}$ & $\mathbf{0.9688_{\pm0.001}}$  \\
		MAEFHRFormer   & $0.7249_{\pm0.042}$ & $0.8595_{\pm0.023}$ & $0.9514_{\pm0.009}$ & $0.7850_{\pm0.005}$ & $0.8894_{\pm0.009}$ \\ 
		MAETransformer & $0.6907_{\pm0.043}$ & $0.8407_{\pm0.026}$ & $0.9376_{\pm0.016}$ & $0.7875_{\pm0.016}$ & $0.8769_{\pm0.009}$ \\
		\bottomrule
	\end{tabular}
	\label{tab:maeeg_baseline_2_class}
\end{table}

\begin{figure}[htbp]
	\centering
	\includegraphics[width=0.45\textwidth]{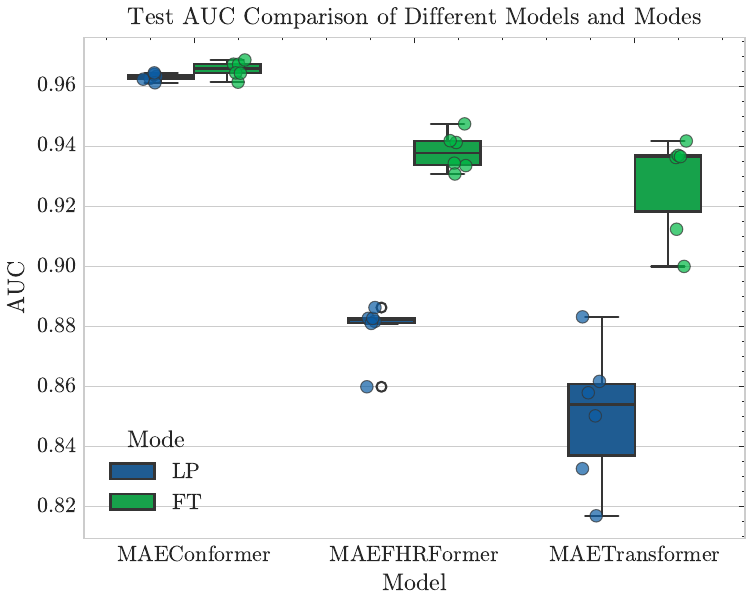}
	\caption{Different MAE-EEG models linear probe (LP) and fine-tune (FT) AUC (4-class over the one-hour epoch-level) distributions from the test set. Each model was evaluated over six random experiments.}
	\label{fig:maeeg_auc_distribution}
\end{figure}

The distribution of test AUC across six independent runs (with different seeds) for three MAE-based models is shown in \cref{fig:maeeg_auc_distribution}. The MAEConformer achieves the highest median (linear probe: 96\%; fine-tune: 97\%) together with the smallest interquartile range (IQR), indicating both superior accuracy and strong robustness to random initialization. In the fine-tune setting, MAEFHRFormer and MAETransformer exhibit comparable median test AUC (approximately 94\%). However, MAETransformer shows a noticeably wider spread, reflecting greater sensitivity to initialization. A similar trend is observed in the linear probe results. MAEFHRFormer maintains relatively stable performance, aside from a few outliers. In contrast, MAETransformer demonstrates the largest variance, with test AUC ranging from 82\% to 88\%, suggesting that its learned representations are less consistent and more sensitive to initialization. Overall, these results indicate that MAEConformer not only achieves higher predictive performance but also produces more stable and reliable representations across repeated experiments.

%%%%%%%%%%%%%%%% Figure: eeg test AUC distribution %%%%%%%%%%%%%%%%%%%%

\paragraph{\underline{HRV classification task:}}
For HRV signal reconstruction, the MAEFHRFormer and MAETransformer exhibited representation collapse and were unable to recover the original NN intervals, even when the focal frequency loss was incorporated. This limitation is likely attributable to the comparatively smaller dataset size and its diversity in the HRV domain, which increase the difficulty of stable reconstruction learning. However, when their decoders were replaced with Conformer blocks and a multi-resolution STFT auxiliary loss was introduced, the models were able to successfully reconstruct the raw NN intervals. Since both baselines share the same encoder design, these modifications effectively unified them into the same MAETransformer architecture. For fair comparison, the adapted MAETransformer was configured with the same architectural parameters as the MAEConformer (HRV) model.

For the supervised learning baselines, only two studies reported classification performance using the same test set. The first one, HRVConformer proposed in \cite{yu2026hrvconformer}, employed the Conformer network trained on a large amount of weakly labelled HRV data. The other one trained and tested on the same datasets as the present study, but instead of using the HRV signal, it used the spectrogram of ECG as input and CNN as backbone to classify HIE severity (ECGSPG-CNN) \cite{rezaeiDeepLearningApproach2025}.

The performance metrics of the different models are summarized in \cref{tab:maehrv_baseline_2_class}. The best overall performance is achieved by the weakly supervised HRVConformer, closely followed by MAEConformer under linear probe setting. However, it is important to note that the HRVConformer requires a substantially larger labelled dataset (1,538 one-hour epochs) and considerably longer supervised training time. In contrast, the linear probe of MAEConformer was only optimized using a simple linear classifier on much smaller labelled dataset (approximately 206 one-hour epochs), yet achieved comparable performance. This clearly demonstrates the strong data efficiency of the pretrained MAEConformer representation. 

Due to limited amount of labelled data, the unfreezing of more encoder layers during fine-tune tends to introduce overfitting, which is reflected by the slight performance drop compared to linear probing. The linear probe performance of MAETransformer is marginally lower than the MAEConformer, while its fine-tune performance is slightly higher. This suggests that with convolution-enhanced Transformer decoder and STFT-based reconstruction loss, the MAETransformer can achieve comparable performance to the MAEConformer after adaptation.

%%%%%%%%%%%%%%% table of HRV baseline comparison %%%%%%%%%%%%%%%%%%%%%%%%%
\begin{table}[htbp]
	\caption{Comparison of different models performance on the HRV(or ECG) test set (epoch-level) with 2-class prediction.}
	\centering
	\begin{tabular}{l|cc|cc}
		\toprule
		& \multicolumn{2}{c|}{fine-tune} & \multicolumn{2}{c}{linear probe} \\
		&    ACC   &   AUC   &    ACC          &     AUC       \\ 
		\midrule
		MAEConformer   & $0.7222_{\pm0.019}$   & $0.8041_{\pm0.014}$ & $\mathbf{0.7395_{\pm0.013}}$ & $\mathbf{0.8242_{\pm0.008}}$       \\
		MAETransformer & $0.7234_{\pm0.019}$ & $0.8113_{\pm0.007}$ & $0.7290_{\pm0.011}$ & $0.8161_{\pm0.007}$  \\ \hline
		HRVConformer \cite{yu2026hrvconformer}   & $\mathbf{0.7479_{\pm0.018}}$   & $\mathbf{0.8323_{\pm0.010}}$ &   -   &    -  \\
		ECGSPG-CNN \cite{rezaeiDeepLearningApproach2025}     & 0.7039   & 0.8083 &     -           &       -        \\
		\bottomrule
	\end{tabular}
	\label{tab:maehrv_baseline_2_class}
\end{table}

\begin{figure}[htbp]
	\centering
	\includegraphics[width=0.45\textwidth]{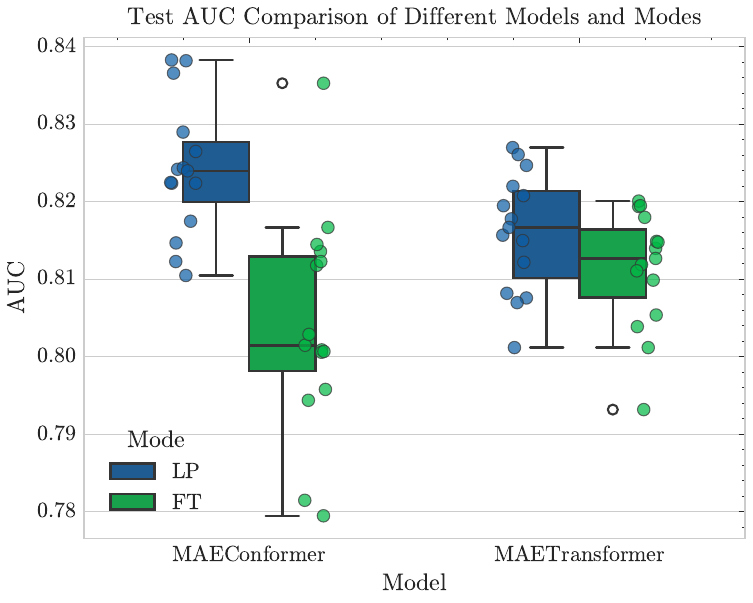}
	\caption{Linear probe (LP) and fine-tune (FT) MAE-HRV models AUC (2-class on the epoch-level) distributions for the test set. Each model was initialized with three random pretrained models and evaluated on five random train/validation data splits.}
	\label{fig:hrv_auc_distribute}
\end{figure}

The test AUC distribution of different MAE models is shown in \cref{fig:hrv_auc_distribute} for HRV. Each model was pretrained using three independent random initializations, and each pretrained checkpoint was subsequently fine-tuned over five different train/validation splits. During fine-tuning, only the top 2 layers of the encoder and classifier head were optimized for the same number of epochs. All evaluations were conducted on the strongly labelled 2-class HRV dataset. 

Under the linear probing setting, the MAEConformer exhibits a higher median AUC and a tighter IQR compared to MAETransformer, indicating more stable linear separable pretrained representations. In contrast, the MAETransformer has slightly higher median and smaller spread under fine-tune. This may suggest MAEConformer is more sensitive to partial parameter unfreezing on a smaller dataset, likely due to its larger effective capacity when additional parameters are optimized.

%%%%%%%%%%%%%%%% Figure: HRV test AUC distribution %%%%%%%%%%%%%%%%%%%%

\subsection{Ablation study}
A series of ablation studies were conducted to evaluate the effect of different configurations on the test performance. The experiments were conducted with the MAE-EEG model. The test performance was reported on the strongly labelled 4-class EEG dataset from ANSeR1, which includes a large number of recordings, providing a reliable evaluation. Each experiment was pretrained for 300 epochs (downstream validation performance plateaued after that shown in \cref{fig:eeg_pretrain_epoch_val_auc}) over the whole pretrained EEG dataset, and linear probing for 200 epochs, and fine-tuning for 500 epochs. 

\paragraph{\underline{Model capacities:}}
The effect of different model configurations on the downstream classification performance is summarized in \cref{tab:model_params_ablation}. It can be seen that different model capacities (model width and depth) only have a minor impact on the linear probing (lp) and fine-tune (ft), indicating a stable performance. It is observed that different decoder depths has relatively large impact on the linear probing performance. A shallower decoder (1-2 layers) shows relatively modest linear probing performance, while their fine-tune performance are not effected. This may suggest weaker pretrained representations can be compensated by fine-tuning.

%%%%%%%%%%%%%%%%%%% Table of model parameters ablation %%%%%%%%%%%%%%%%%%%%%%%%
\begin{table}[htb]
	\centering
	\caption[MAEConformer (EEG) model configuration ablation experiments]{MAEConformer (EEG) model configuration ablation experiments evaluated on 4-class EEG dataset. Experiments default setting: 10\,s patch size, mask ratio 0.4 and Conformer decoder. The default settings are marked with \colorbox{gray!30}{grey}.} % Main table caption
	\label{tab:model_params_ablation}
	
	\begin{minipage}{\shorttablewidth}
		\centering
		\captionsetup{labelformat=empty} % Remove automatic numbering
		\caption*{(a) Encoder layers} % Manually set caption
		\begin{tabular}{lcc}
			\toprule
			&     ft  &    lp	      \\
			\midrule
			3 		  &   0.9630      &   0.9544     \\
			4         &   0.9654      &   0.9614    \\
			5         &   0.9653      &   0.9598    \\
			6         &   \colorbox{gray!30}{0.9646}  &  \colorbox{gray!30}{0.9603}     \\  
			7         &   \textbf{0.9689}    &   \textbf{0.9647}    \\
			\bottomrule
		\end{tabular}
	\end{minipage} %
	\hfil
	\begin{minipage}{\shorttablewidth}
		\centering
		\captionsetup{labelformat=empty}
		\caption*{(b) Decoder depth} % Manually set caption
		\begin{tabular}{lcc}
			\toprule
			&     ft  &    lp	      \\
			\midrule
			1 		  &   \textbf{0.9655}      &  0.9454     \\
			2 		  &   0.9640      &  0.9530     \\
			3 		  &   \colorbox{gray!30}{\textbf{0.9646}}  &  \colorbox{gray!30}{0.9603}     \\
			4         &   0.9584      &  \textbf{0.9651}     \\
			5         &   0.9639      &  0.9598     \\
			\bottomrule
		\end{tabular}
	\end{minipage}  %
	\hfil
%	\bigskip % Space between rows
	\begin{minipage}{\shorttablewidth}
		\centering
		\captionsetup{labelformat=empty}
		\caption*{(c) encoder dimension}
		\begin{tabular}{lcc}
			\toprule
			&     ft  &    lp	      \\
			\midrule
			128       &   \textbf{0.9670}      &  0.9522    \\
			144       &   0.9629      &  \textbf{0.9646}     \\
			256       &   \colorbox{gray!30}{0.9646}  &  \colorbox{gray!30}{0.9603}     \\
			512       &   0.9648      &  0.9626     \\
			\bottomrule
		\end{tabular}
	\end{minipage} %
	\hfil
	\begin{minipage}{\shorttablewidth}
		\centering
		\captionsetup{labelformat=empty}
		\caption*{(d) patch size}
		\begin{tabular}{lcc}
			\toprule
			&     ft        &    lp	      \\
			\midrule
			5s    &   0.9621      &    0.9578  \\
			10s   &   \colorbox{gray!30}{\textbf{0.9646}}  &  \colorbox{gray!30}{\textbf{0.9603}}     \\
			15s   &   0.9473      &    0.8918   \\
			20s   &   0.9628      &    0.9342   \\
			\bottomrule
		\end{tabular}
	\end{minipage} %
	\hfil
%	\bigskip % Space between rows
	\begin{minipage}{\shorttablewidth}
		\centering
		\captionsetup{labelformat=empty}
		\caption*{(e) decoder dimension}
		\begin{tabular}{lcc}
			\toprule
			&     ft &    lp	      \\
			\midrule
			64       &   0.9628      &  0.9638    \\
			128      &   \colorbox{gray!30}{\textbf{0.9646}}  &  \colorbox{gray!30}{0.9603}     \\
			256      &   \textbf{0.9648}      &  \textbf{0.9670}     \\
			\bottomrule
		\end{tabular}
	\end{minipage}%
	\hfil %
	\begin{minipage}{\shorttablewidth}
		\centering
		\captionsetup{labelformat=empty}
		\caption*{(f) loss functions}
		\begin{tabular}{lcc}
			\toprule
			&     ft     &    lp	      \\
			\midrule
			full 		 &   \colorbox{gray!30}{\textbf{0.9646}}  &  \colorbox{gray!30}{0.9603}     \\
			w/o mrstft   &   0.9456      &  0.8753     \\
			w/o unif     &   0.9615      &  \textbf{0.9621}     \\
			\bottomrule
		\end{tabular}
	\end{minipage}%
%	\bigskip % Space between rows
\end{table}

A patch size of 10\,s appears to represent an optimal balance between local and global temporal modelling, yielding the highest performance in both linear probing and fine-tuning evaluations. A pronounced performance degradation is observed when larger patch sizes are employed, indicating that overly coarse tokenization may obscure important short-term physiological dynamics. In comparison, the uniformity loss has only a minor influence on downstream classification performance. Conversely, the removal of the multi-resolution STFT loss results in a substantial decline in linear probing performance, underscoring the importance of frequency-domain supervision for learning robust and transferable representations during self-supervised pretraining.

\paragraph{\underline{Decoder type:}} The MAE-EEG as default employed the Conformer decoder, which has been replaced with Transformer and cross-attention decoder (keeping the same depth and embedding dimension) adapted from MAETransformer and MAEFHRFormer (the encoder remains the Conformer). The Cross-Attention and Conformer decoder show minor differences on the linear probing and fine-tune performance. A significant performance drop is observed from the linear probing when the decoder is changed to a Transformer (shown in \cref{tab:model_architecture_ablation}), even through it was compensated for to fine-tune with Conformer encoder. This suggests that plain global attention is insufficient to help the model to learn meaningful representations.

%%%%%%%%%%%%%%%%%%% Table of architectural ablation %%%%%%%%%%%%%%%%%%%%%%%%
\begin{table}[htb]
	\centering
	\caption[MAEConformer (EEG) architectural configuration ablation experiments]{MAEConformer (EEG) architectural configuration ablation experiments evaluated on 4-class EEG dataset. The default settings are marked with \colorbox{gray!30}{grey}.} % Main table caption
	\label{tab:model_architecture_ablation}
	\begin{minipage}{0.4\textwidth}
		\centering
		\captionsetup{labelformat=empty}
		\caption*{(a) position embedding}
		\begin{tabular}{lcc}
			\toprule
			&     ft     &    lp     \\
			\midrule
			relative       &   \colorbox{gray!30}{\textbf{0.9646}}   &  \colorbox{gray!30}{0.9603}    \\
			absolute       &   0.9553   &  \textbf{0.9647}    \\
			\bottomrule
		\end{tabular}
	\end{minipage}%
	\hfill
	\begin{minipage}{0.4\textwidth}
		\centering
		\captionsetup{labelformat=empty}
		\caption*{(b) GRN vs BN}
		\begin{tabular}{lcc}
			\toprule
			&     ft     &    lp     \\
			\midrule
			GRN       &   \colorbox{gray!30}{\textbf{0.9646}}   &  \colorbox{gray!30}{0.9603}   \\
			BN        &   0.9618    &  \textbf{0.9653}     \\
			\bottomrule
		\end{tabular}
	\end{minipage}%
%	\bigskip % Space between rows
	\hfill
	\begin{minipage}{0.4\textwidth}
		\centering
		\captionsetup{labelformat=empty}
		\caption*{(c) decoder type}
		\begin{tabular}{lcc}
			\toprule
			&     ft     &    lp     \\
			\midrule
			Conformer       &   \colorbox{gray!30}{0.9646} &   \colorbox{gray!30}{0.9603}    \\
			Transformer     &   0.9504   &   0.8266   \\
			CrossAttn       &   \textbf{0.9687}   &   \textbf{0.9610}   \\
			\bottomrule
		\end{tabular}
	\end{minipage}%
	\hfill
	\begin{minipage}{0.4\textwidth}
		\centering
		\captionsetup{labelformat=empty}
		\caption*{(d) classifier head}
		\label{subtab:ablat_heads}
		\begin{tabular}{lc}
			\toprule
			&     ft          \\
			\midrule
			mlp               &  0.9650       \\
			fcn               &  \colorbox{gray!30}{0.9646}   \\
			linear            &  \textbf{0.9672}       \\
			mlp\_cls          &  0.9635       \\
			mlp\_global\_pool &  0.9663       \\
			\bottomrule
		\end{tabular}
	\end{minipage}%
	
\end{table}

\paragraph{\underline{GRN vs BN:}}
The GRN layer in the convolutional module is designed to promote feature diversity by enhancing inter-channel competition. To evaluate its contribution to downstream classification performance, the GRN was replaced with a conventional Batch Normalization layer. Although the overall performance difference is modest, distinct patterns emerge. The model with GRN exhibits slightly lower linear probing performance, but achieved improved test AUC after fine-tune. In contrast, the batch normalization yields higher linear probing AUC, yet its fine-tune performance deteriorates. This degradation may be attributed to overfitting of batch-wise statistical estimation on the smaller labelled dataset, leading to reduced generalization ability after parameter unfreezing.

\paragraph{\underline{Position embeddings:}}
The different positional embedding schemes were also studied. The default learnable relative positional embeddings (\textit{T5-style} \cite{raffelExploringLimitsTransfer2019}) used in both encoder and decoder were replaced by fixed absolute sinusoidal (\textit{sin-cos}) positional embeddings. Under the linear probing setting, both schemes demonstrate minimal differences, with the absolute position embedding achieving marginally higher performance. However, when the model is fully tuned, the performance of the absolute position embedding declines. This suggests that the strongly encoded absolute positional information may constrain representation flexibility during full-network adaptation, whereas relative positional encoding provides greater robustness when fine-tuning on limited labelled data.

\paragraph{\underline{Classifier heads:}}
The impact of different classifier heads on the fine-tune performance were also investigated. The \textit{mlp} head first flattens the entire sequence embedding (excluding the class token) and projects it to a hidden dimension of 512. After a non-linear activation and dropout layer, the hidden representation is linearly projected to a number of output classes. The \textit{fully convolutional network (fcn)} head takes the encoder output as input. After removing the class token, the patch embeddings are fed into a convolutional layer that projects the sequence dimension to the number of classes. The convolutional filters operate along the embedding dimension with large kernel size (7) and stride of 2. Following a non-linear activation and batch normalization, a second convolutional layer (kernel size 3, stride 2) maintains the channel dimension equal to the number of classes, while further reduce the feature map size. An average pooling followed by global pooling progressively reduces the spatial dimension to 1, and final predictions were obtained from the channel dimension. 

The \textit{linear classifier} head adopted the same architecture as the linear probing setup but incorporates a learnable affine transformation within the batch normalization layer. The sequence embeddings, except for the class token patches, are flattened, normalized, and then linearly project to the class dimension. The \textit{mlp\_global\_pool} head averages embeddings across all patches (without considering the class token). The resultant representation is passed through an MLP block with two hidden layer (hidden dimension equal to the encoder dimension and output class) to produce predictions. In contrast, the \textit{mlp\_cls} head relies solely on the class token embedding and applied the same MLP block -- this will evaluate whether the class token has effectively aggregated global contextual information from the sequence. 

The fine-tune results shown in \cref{subtab:ablat_heads} demonstrate consistently stable performance across different classifier heads, suggesting that a choice of head architecture has a limited impact once strong pretrained representations are available. Notably, even the fcn head which contains substantially fewer parameters than the other heads, achieves a comparable performance. Furthermore, the minor difference between class token based head and the global average pooling head indicates that the class token has successfully captured and integrated information from the entire sequence.

\section{Discussion}
\subsection{Attention analysis}
Although masked autoencoder based models have demonstrated strong performance in representation learning for physiological time-series, their internal attention decision mechanisms remain poorly understood. In the absence of supervision from downstream tasks, models are optimized to reconstruct missing inputs. It is possible that models with similar losses or downstream metrics can learn fundamentally different representations. Consequently, it is necessary to evaluate the model from the representation quality perspective, which the performance metrics may not reveal. 

To better understand how architectural choices shape representation structure, attention entropy and attention distance are analysed as task-independent diagnostics of attention behaviour. Attention entropy quantifies the degree of head-wise selectivity and diversity, revealing potential capacity collapse or redundancy, while attention distance characterizes the temporal scope over which information is integrated. By comparing these metrics across different architectures, we aim to elucidate how inductive bias influences the attention structure, robustness, and the ability to capture multi-scale temporal dependencies in physiological signals.

\begin{figure}[htbp]
	\centering
	\begin{subfigure}[b]{0.3\textwidth}
		\centering
		\includegraphics[width=\textwidth]{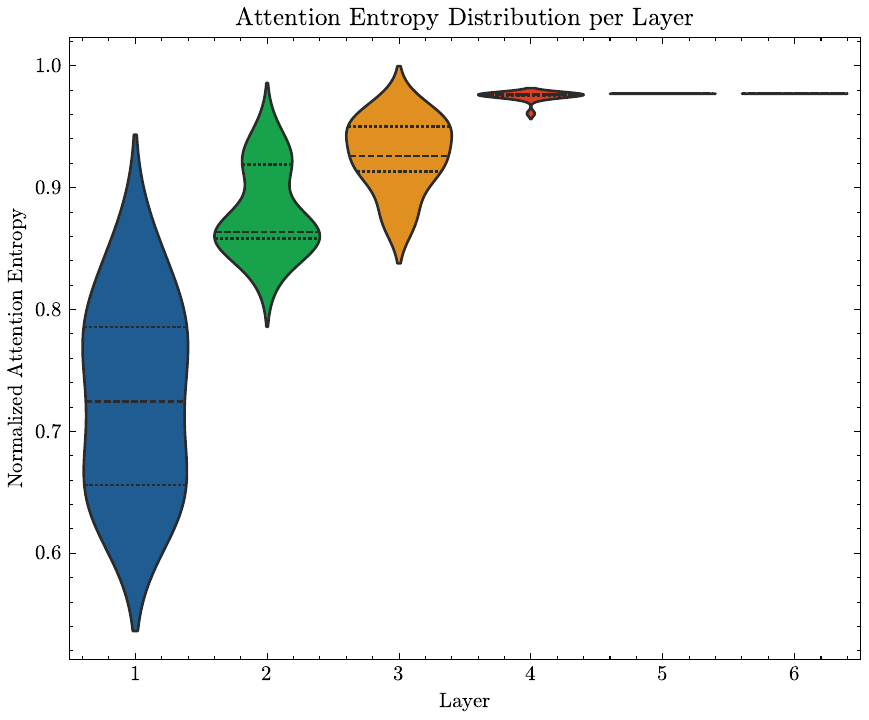}
		\caption{MAETransformer}
		\label{subfig:attn_violin_maetransformer}
	\end{subfigure}
	\hfill
	\begin{subfigure}[b]{0.3\textwidth}
		\centering
		\includegraphics[width=\textwidth]{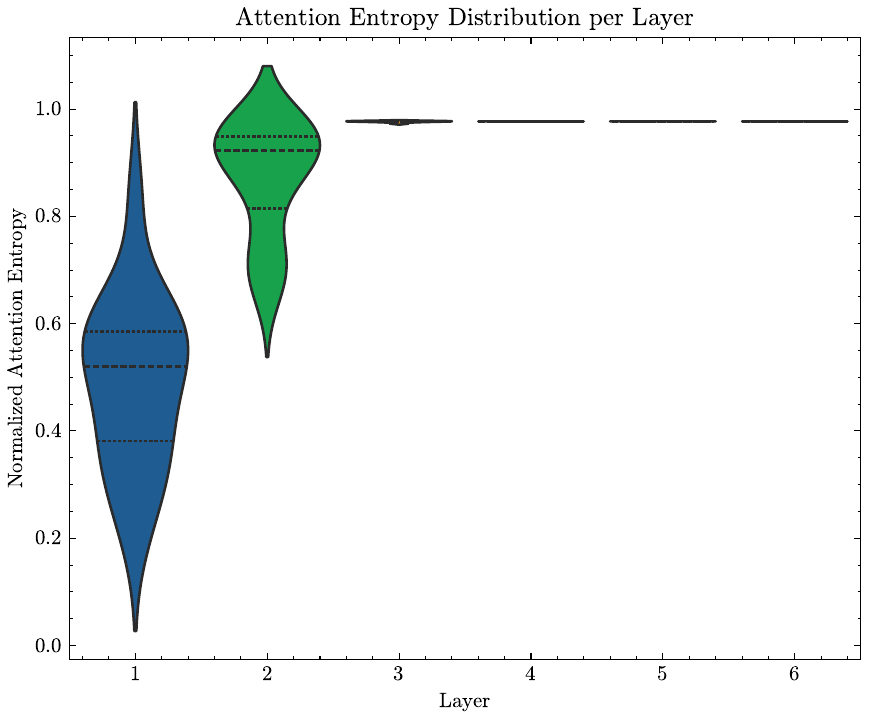}
		\caption{MAEFHRFormer}
		\label{subfig:attn_violin_maefhrformer}
	\end{subfigure}
	\hfill
	\begin{subfigure}[b]{0.3\textwidth}
		\centering
		\includegraphics[width=\textwidth]{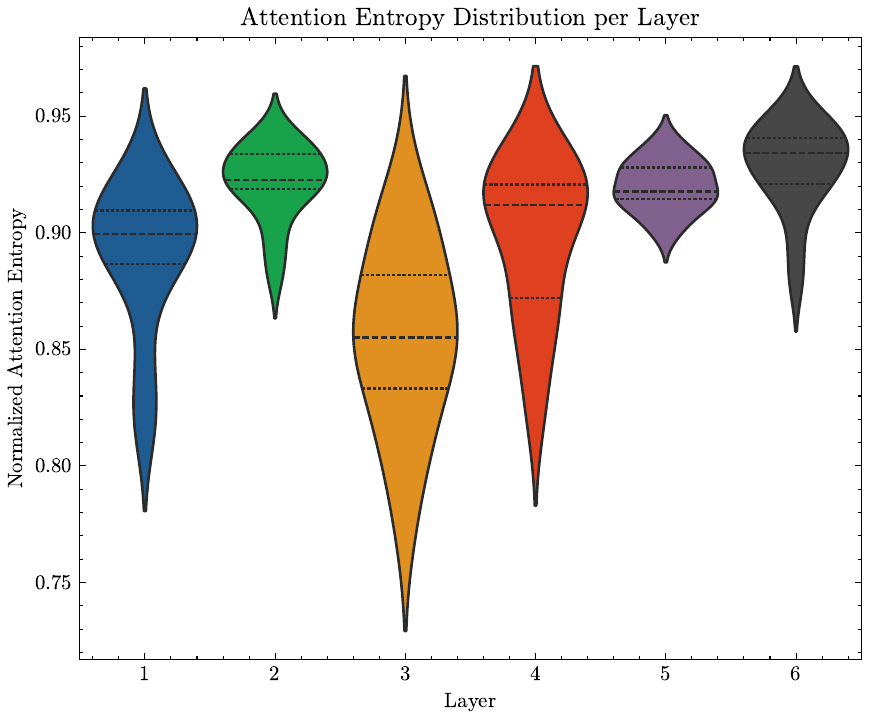}
		\caption{MAEConformer}
		\label{subfig:attn_viloin_maeconformer}
	\end{subfigure}
	\caption{Average attention entropy distribution across all test samples of each layer and head from the encoder of the different models. Entropy has been normalized to [0,1] (values out of this range do not indicate a realistic value number as they are caused by the Kernel Density Estimation using in the violin plot). The encoders from these three MAE-EEG models have the same number of attention heads and same sequence length.}
	\label{fig:attn_entropy_violin_plots}
\end{figure}

Attention entropy from the MAE-EEG model is averaged across all test samples to get a stable estimation. To better compare these distributions, attention entropies have been normalized by their maximum value $log(L)$ to the range of [0,1]. \cref{fig:attn_entropy_violin_plots} demonstrates head-wise attention entropy distributions across encoder layer and head using violin plots. The width of the violin indicate the density of heads at that entropy value. The median and Interquartile range (IQR) lines are also marked in each violin. Healthy attention behaviour is characterized by entropy values within the interior of the range and a non-zero IQR value, indicating effective utilization of attention capacity. In contrast, low-entropy collapse corresponds to near-deterministic attention across many heads, while high-entropy uniformization reflects degenerate uniform attention with diminished head diversity.

%%%%%%%%%%%%%%%% Figure: attention entropy violin %%%%%%%%%%%%%%%%%%%%

Representative runs from each model were selected based on stable, non-pathological entropy patterns. As shown in \cref{fig:attn_entropy_violin_plots} both the MAETransformer and the MAEFHRFormer exhibit a pronounced tendency toward uniform attention in deeper layers, where head-wise entropy rapidly saturates near its theoretical maximum and inter-head variability diminishes. In contrast, the MAEConformer maintains non-trivial entropy level and preserves head diversity across all layers, indicating more heterogeneous and information-rich attention patterns. Furthermore, across multiple random seeds, the MAETransformer and MAEFHRFormer frequently display early-layers collapse, manifested as extremely low entropy in several attention heads within the same layer. These observations suggest the proposed method with convolution-attention coupling architecture provides a structural advantage in preventing degenerate attention behaviour. Compared with purely global attention or cross-attention based designs, MAEConformer better preserves head diversity and more effectively utilizes its attention capacity, leading to richer and more stable internal representations.

\begin{figure}[htbp]
	\centering
	\begin{subfigure}[b]{0.3\textwidth}
		\centering
		\includegraphics[width=\textwidth]{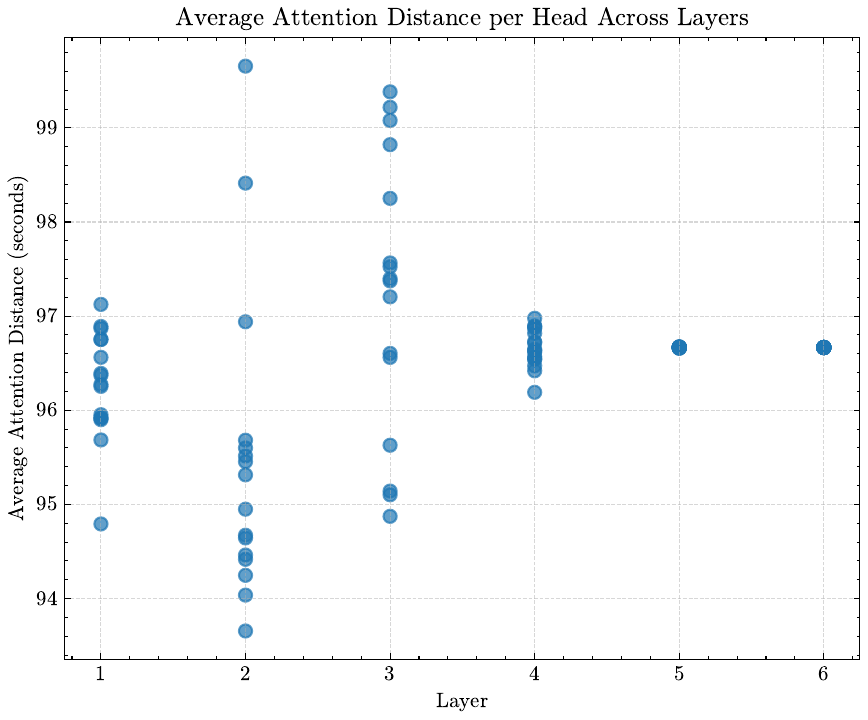}
		\caption{MAETransformer}
		\label{subfig:attn_dis_maetransformer}
	\end{subfigure}
	\hfill
	\begin{subfigure}[b]{0.3\textwidth}
		\centering
		\includegraphics[width=\textwidth]{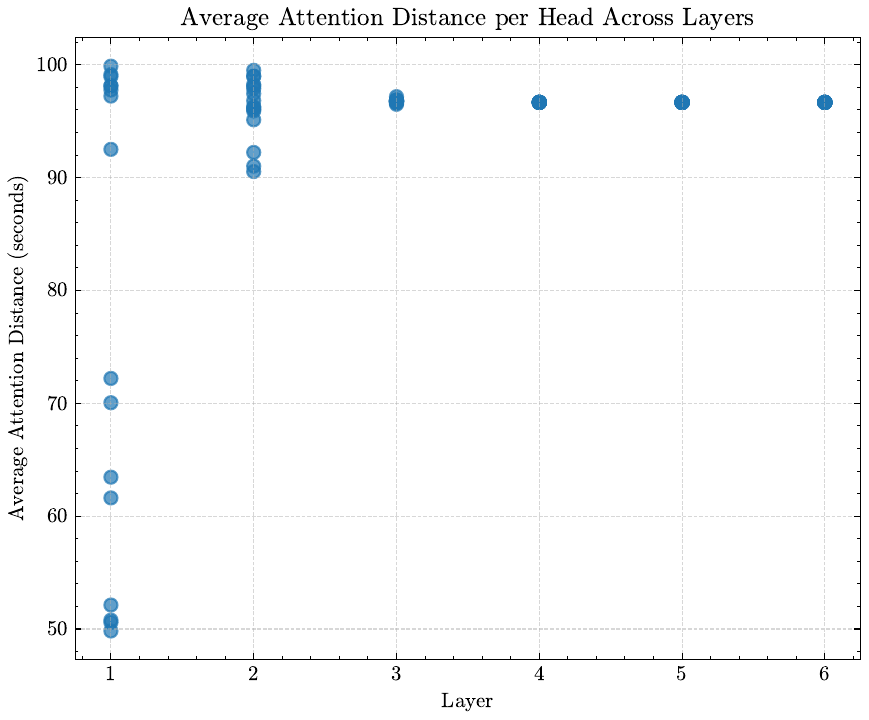}
		\caption{MAEFHRFormer}
		\label{subfig:attn_dist_maefhrformer}
	\end{subfigure}
	\hfill
	\begin{subfigure}[b]{0.3\textwidth}
		\centering
		\includegraphics[width=\textwidth]{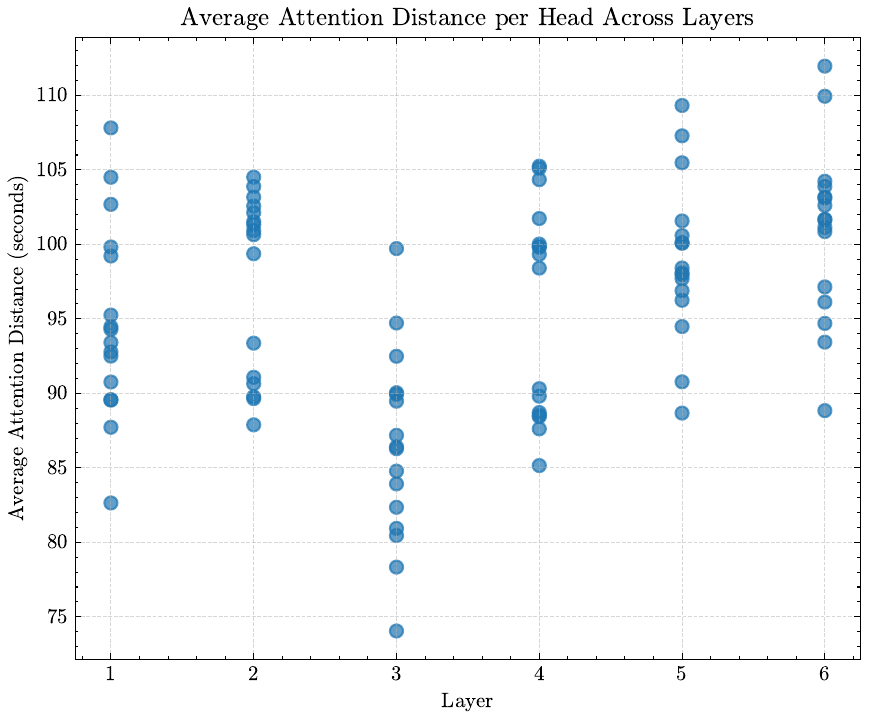}
		\caption{MAEConformer}
		\label{subfig:attn_dist_maeconformer}
	\end{subfigure}
	\caption{Attention distance (in seconds) averaged over the whole test set for each head across layers from different MAE-EEG models. \textbf{(a)} MAETRansformer \textbf{(b)} MAEFHRFormer and \textbf{(c)} MAEConformer.}
	\label{fig:attn_dist_plot}
\end{figure}

Attention distance was computed as the expected relative token distance weighted by the attention matrix. The relative distance is defined as the absolute difference between patch indices. The class token has been removed to make sure each token has a physical position in time. \cref{fig:attn_dist_plot} demonstrates the average attention distance of each head across layers from the encoder of the three pretrained model. Each head attention distance is averaged over all test samples and converted to the time domain (in seconds).

%%%%%%%%%%%%%%%% Figure: attention distance visualization %%%%%%%%%%%%%%%%%%%%

From \cref{fig:attn_dist_plot}, most of the heads from  MAETransformer and MAEFHRFormer integrate information within the range of 90--100\,s, except for the first layer of MAEFHRFormer, few heads of which attend to 50--70\,s distance. In contrast, the MAEConformer integrates information in a broader range (75--110\,s), indicating its capability to capture multiscale temporal dependencies. On the other hand, heads in the later layers from MAETransformer and MAEFHRFormer almost overlapped into very similar attention distance, while heads from MAConformer show a large spread across all layers. This again suggests the richness and non-redundancy of the learned representation.

\subsection{UMAP visualization}
UMAP is a dimension reduction tool, that was employed to visualize how pretrained and fine-tuned features are distributed in the embedding space. Class tokens from the final encoder layer were used as the latent representation of each 5-min window. As adjacent windows are highly correlated because of overlapping, features were L2-normalized and aggregated to the one-hour epoch-level using robust mean pooling; this reduces the sensitivity to amplitude variations and artefact-contaminated windows. Feature distances of window embeddings to their mean were computed, with the top/bottom 10\% windows removed to ensure consistent latent structure. UMAP reduces the feature dimension from the encoder embedding dimension to two for the convenience here of visualization.

\begin{figure}[htb]
	\centering
	\begin{subfigure}[b]{0.4\textwidth}
		\centering
		\includegraphics[width=\textwidth]{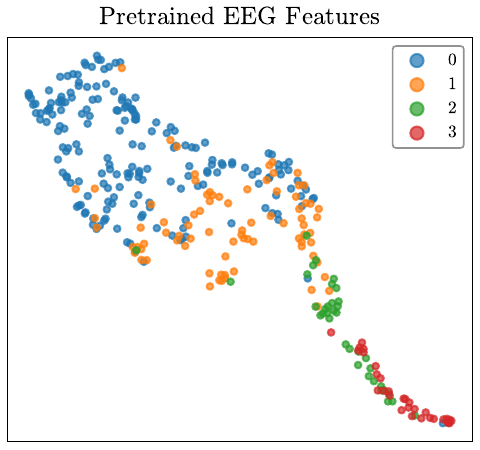}
		\caption{}
		\label{subfig:umap_eeg_prt}
	\end{subfigure}
	\hfill
	\begin{subfigure}[b]{0.4\textwidth}
		\centering
		\includegraphics[width=\textwidth]{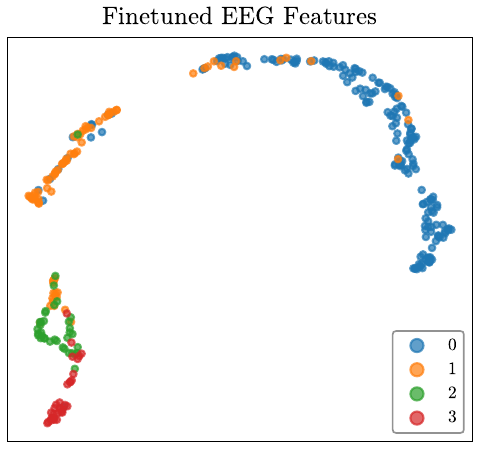}
		\caption{}
		\label{subfig:umap_eeg_ft}
	\end{subfigure}
	\caption{UMAP visualization for the \textbf{(a)} pretrained and \textbf{(b)} fine-tuned Conformer encoder (EEG). Each point denotes the UMAP reduced embeddings of the encoder output from the strongly labelled one-hour ANSeR1 EEG recording. The 4-class labels are marked with different colours.}
	\label{fig:eeg_umap_visualization}
\end{figure}

\cref{fig:eeg_umap_visualization} demonstrates the UMAP embeddings from the strongly labelled EEG test set. The pretrained model shows emergent clustering of HIE severity levels without label supervision, suggesting that the model is capable of learning meaningful representations with the masking and reconstruction objective. After fine-tuning, the different classes show clearer separation, indicating that the fine-tuning refined the representation and improved task-specific discrimination. 

%%%%%%%%%%%%%%%% Figure: UMAP EEG visualization %%%%%%%%%%%%%%%%%%%%

\begin{figure}[htb]
	\centering
	\begin{subfigure}[b]{0.4\textwidth}
		\centering
		\includegraphics[width=\textwidth]{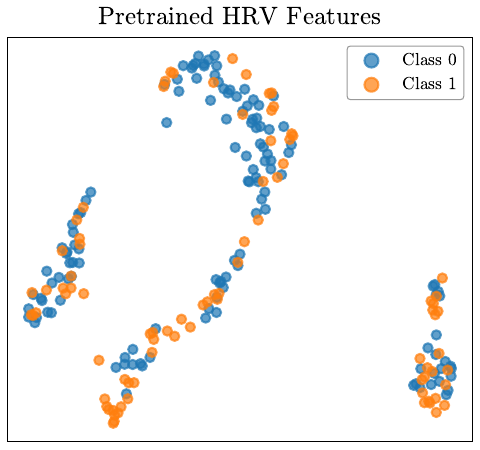}
		\caption{}
		\label{subfig:umap_hrv_prt}
	\end{subfigure}
	\hfill
	\begin{subfigure}[b]{0.4\textwidth}
		\centering
		\includegraphics[width=\textwidth]{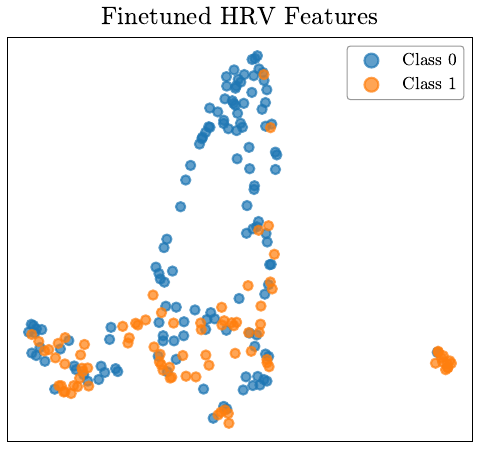}
		\caption{}
		\label{subfig:umap_hrv_ft}
	\end{subfigure}
	\caption{UMAP visualization for the \textbf{(a)} pretrained and \textbf{(b)} fine-tuned Conformer encoder (HRV). Each point denotes the UMAP reduced embeddings of the encoder output from the strongly labelled one-hour ANSeR1 HRV recording. The 2-class labels are marked with different colours.}
	\label{fig:hrv_umap_visualization}
\end{figure}

The pretrain and fine-tune model features from the strongly labelled HRV test set is shown in \cref{fig:hrv_umap_visualization}. The pretrained representation shows greater overlapping than the fintune in 2D visualization, reflecting the preservation of intra-class diversity. After fine-tuning, the features become more class-aligned, resulting in tighter clustering. However, the linear probing performance indicates that the pretrained embeddings maintain strong linear separability from the original high-dimension space, suggesting that the 2D visualization does not fully reflect the discriminative capacity.

%%%%%%%%%%%%%%%% Figure: UMAP HRV visualization %%%%%%%%%%%%%%%%%%%%

\subsection{Clinical alignment}
To demonstrate model confidence and compare prediction dynamics across modalities, epoch-level prediction probability derived from the EEG and the HRV signal is visualized over time. To avoid confusion, the binary (2-class) models are used here, while weak labels and strongly labelled expert grades (5-grade scale: 0--4) were overlaid for reference. Recordings are selected from the ANSeR1 test set within the postnatal ages between range of 6 to 48 hours (where applicable) to facilitate temporal alignment with expert assessments. Window-level prediction probabilities were averaged to obtain epoch-level estimates, and the corresponding standard deviation was computed to provide an indication of the prediction uncertainty.

\begin{figure}[htb]
	\centering
	\begin{subfigure}[b]{0.8\textwidth}
		\centering
		\includegraphics[width=\textwidth]{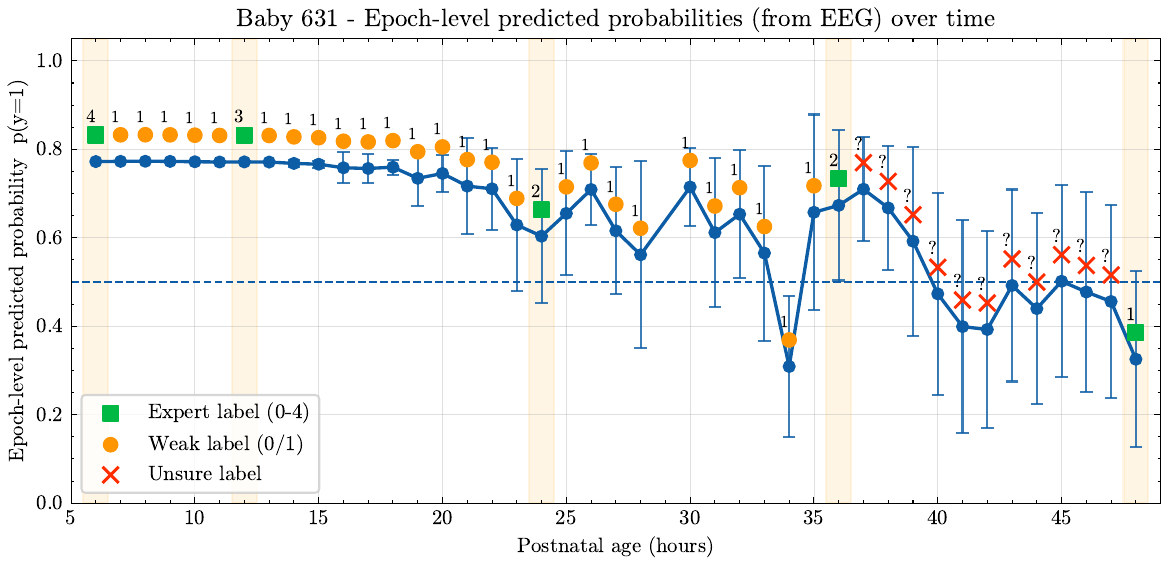}
		\caption{}
		\label{subfig:epoch_probs_eeg}
	\end{subfigure}
	\hfill
	\begin{subfigure}[b]{0.8\textwidth}
		\centering
		\includegraphics[width=\textwidth]{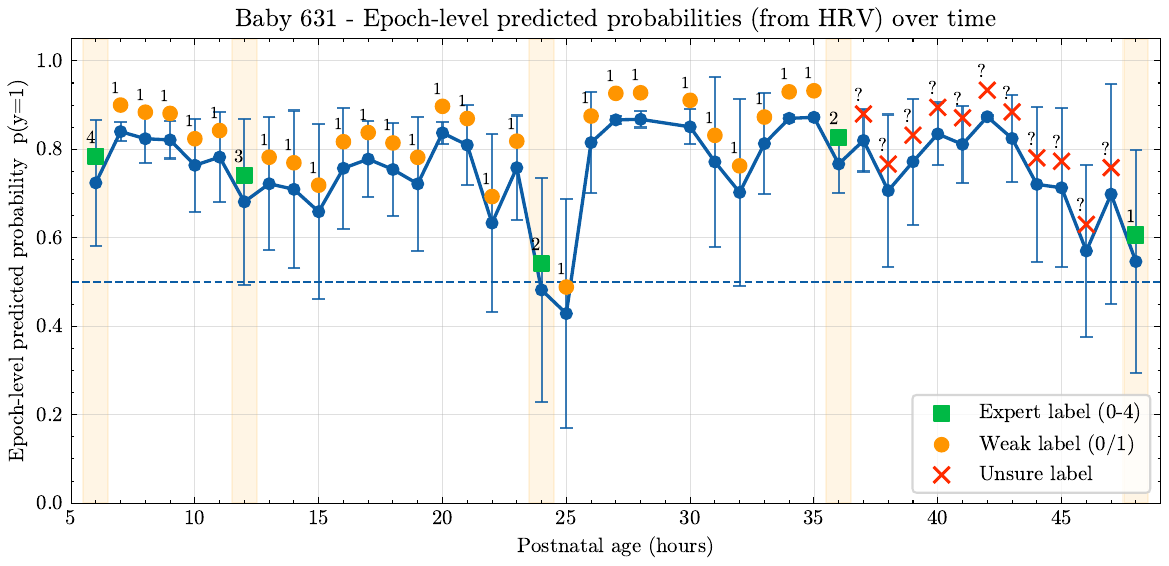}
		\caption{}
		\label{subfig:epoch_probs_hrv}
	\end{subfigure}
	\caption[Example of model prediction probabilities over postnatal ages from the ANSeR1 EEG and HRV signal]{Example of model prediction probabilities over postnatal ages from the ANSeR1 \textbf{(a)} EEG and \textbf{(b)} HRV signal. Each point indicates  the prediction probability of the unhealthy (moderate-to-severe HIE, corresponding to the expert label: 2-4) class p(y=1) of a single one-hour epoch. The unsure label indicates the weak label cannot be determined because the expert labels change between grade 1 and 2 (the boundary of mild and moderate-to-severe). The error bar represents the standard deviation over all the  prediction probabilities within the one-hour window.}
	\label{fig:epoch_probs_plots}
\end{figure}

An example of the temporal evolution of prediction probabilities from both EEG and HRV models is shown in \cref{fig:epoch_probs_plots}. The EEG model exhibits more stable predictions, with smoother probability trajectories and generally smaller within-epoch standard deviations, whereas the HRV model shows greater variability. This difference is consistent with EEG directly reflecting cortical activity, while HRV provides a more indirect measure of neurological injury and may be influenced by broader autonomic and physiological factors. Although label smoothing (0.1 for both models) may discourage overly confident predictions, the absolute probability scales remain model- and modality-dependent. Therefore, probabilities around 0.8 represent relatively strong positive evidence within the observed range of the EEG model and should not be directly equated with the same probability values produced by the HRV model.

Interestingly, both curves exhibit a similar tendency -- when the expert labels reduce, the prediction probability of class 1 gradually declines (particularly for the EEG model). This means that the probability behaves like an ordinal severity estimator, suggesting the representation preserves disease progression structure, even through it was trained as a binary classification task. 

%%%%%%%%%%%%%%%% Figure: UMAP HRV visualization %%%%%%%%%%%%%%%%%%%%

In addition, the consistent temporal trajectory across EEG and HRV suggests that the models capture a common pathophysiological evolution rather than modality-specific artefacts. The severity evolution is reflected both on cortical (EEG) activity and the autonomic (HRV) system, suggesting that the model has learned physiologically meaningful features. Notably, the HRV reaction seems to slightly lag behind, which aligns with the hypothesis that HRV is an indirect indicator of HIE severity.

\section{Conclusion}
In this paper, we proposed the MAEConformer, a unified self-supervised learning framework that combines the Conformer architecture with the Masked Autoencoder (MAE) paradigm for large-scale pretraining on unlabelled EEG and HRV data. The framework learns task-agnostic representations by reconstructing masked physiological signals and incorporates a multi-resolution short-time Fourier transform (MR-STFT) loss to enhance spectral representation learning.

The proposed approach demonstrated strong transferability and data efficiency on downstream HIE severity classification tasks. In particular, MAE-EEG achieved state-of-the-art test AUCs of 97.19\% and 96.56\% for binary and four-class EEG-based HIE severity classification, respectively, outperforming both self-supervised and supervised baselines. For the HRV-based HIE classification task, the MAE-HRV model achieved 82.42\% test AUCs surpassing the self-supervised Transformer and supervised convolution baselines. Additional analyses further showed that the learned representations capture meaningful disease-related structures while reducing feature redundancy.

These findings highlight the potential of large-scale self-supervised pretraining to address the scarcity of expert annotations in neonatal neurocritical care. More broadly, MAEConformer provides a scalable foundation for representation learning from physiological time series and may facilitate the development of more robust, data-efficient, and clinically useful decision-support systems.

\bibliographystyle{unsrtnat}
\bibliography{references}  %%% Uncomment this line and comment out the ``thebibliography'' section below to use the external .bib file (using bibtex) .

%%% Uncomment this section and comment out the \bibliography{references} line above to use inline references.
% \begin{thebibliography}{1}

% 	\bibitem{kour2014real}
% 	George Kour and Raid Saabne.
% 	\newblock Real-time segmentation of on-line handwritten arabic script.
% 	\newblock In {\em Frontiers in Handwriting Recognition (ICFHR), 2014 14th
% 			International Conference on}, pages 417--422. IEEE, 2014.

% 	\bibitem{kour2014fast}
% 	George Kour and Raid Saabne.
% 	\newblock Fast classification of handwritten on-line arabic characters.
% 	\newblock In {\em Soft Computing and Pattern Recognition (SoCPaR), 2014 6th
% 			International Conference of}, pages 312--318. IEEE, 2014.

% 	\bibitem{hadash2018estimate}
% 	Guy Hadash, Einat Kermany, Boaz Carmeli, Ofer Lavi, George Kour, and Alon
% 	Jacovi.
% 	\newblock Estimate and replace: A novel approach to integrating deep neural
% 	networks with existing applications.
% 	\newblock {\em arXiv preprint arXiv:1804.09028}, 2018.

% \end{thebibliography}

\end{document}